\documentclass[10pt,twocolumn,letterpaper]{article}

\usepackage{color}
\usepackage{booktabs}
\usepackage{iccv}
\usepackage{epsfig}
 \pdfoutput=1
\usepackage{graphicx}
\usepackage{amsmath}
\usepackage{amssymb}
\usepackage{setspace}
\usepackage{multirow}
\usepackage{soul}
\usepackage{rotating}
\usepackage{caption}
\usepackage{bbold}
\usepackage{float}
\usepackage{subfig}
\usepackage{algorithm}
\usepackage{verbatim}
\usepackage[noend]{algpseudocode}
\usepackage{cite}
\usepackage{afterpage}
\usepackage{textgreek}
\usepackage{chemgreek}
\usepackage{fp}
\usepackage[outline]{contour}
\contourlength{1pt}
\usepackage{enumitem}
\usepackage{greekctr}
\usepackage[pagebackref=true,breaklinks=true,letterpaper=true,colorlinks,bookmarks=false]{hyperref}

\iccvfinalcopy % *** Uncomment this line for the final submission

\def\iccvPaperID{****} % *** Enter the ICCV Paper ID here

\begin{document}

\title{A Study of Cross-domain Generative Models \\ Applied to Cartoon Series}

%%%%%%%%% TITLE

\author{Eman T. Hassan \hspace{2in} David J. Crandall \\
School of Informatics, Computing, and Engineering \\
Indiana University \\
Bloomington, IN USA \\
Email: \{emhassan, djcran\}@indiana.edu}

\maketitle

\begin{abstract}
	We investigate  Generative Adversarial
	Networks (GANs) to model one particular kind of image: frames
	from TV cartoons. Cartoons are particularly interesting because
	their visual appearance emphasizes the important semantic information
	about a scene while abstracting out the less important details, but
	each cartoon series has a distinctive artistic style that performs
	this abstraction in different ways.  We consider a  dataset
	consisting of images from two popular television cartoon series, Family Guy and
	The Simpsons.  We examine the ability of GANs to generate images from
	each of these two domains, when trained independently as well as on
	both domains jointly. We find that generative models may be capable of
	finding semantic-level correspondences between these two image domains despite 
	the
	unsupervised setting, even when the training data
	does not give labeled alignments between them. 
	% We show that the
	%problem is many-to-many mapping and more work to be done to explore
	%this problem.
\end{abstract}

\section{Introduction}
\label{sec:Intro}

Filmmakers and authors may not wish to admit it, but almost every story -- and almost every work of
literature and art in general -- borrows heavily from those that came
before it.  Sometimes this is explicit: the 2004 movie \textit{Phantom
	of the Opera} is a film remake of the famous Andrew Lloyd Webber
musical, which was based on an earlier 1976 musical, which in turn was
inspired by the 1925 silent film, all of which are based on the 1910
novel by Gaston Leroux.  Sometimes the borrowing is for humor -- the
TV sitcom \textit{Modern Family}'s episode \textit{Fulgencio} was a clear
parody of \textit{The Godfather}, for example -- or for political
expression, such as \textit{The Onion}'s satirized version of news
stories. Even highly original stories still share common themes and
ingredients, like archetypes for characters~\cite{archetypes} (``the tragic hero,'' ``the
femme fatal,'' etc.) and plot lines~\cite{sevenplots} (``rags to riches,'' ``the quest,'' etc.).

\begin{figure}[t]
	\centering
	{\footnotesize{\textsf{
				\begin{tabular}{@{}p{4.8cm}@{\,\,}p{0.9cm}@{}p{2cm}@{}}
					\multicolumn{1}{c}{{\textbf{Training (actual) frames}}} & & \multicolumn{1}{c}{\textbf{\,Novel frames}} \\
					\begin{flushleft}
						\vspace{-15pt}
						{\includegraphics[width=1.1cm,height=8mm]{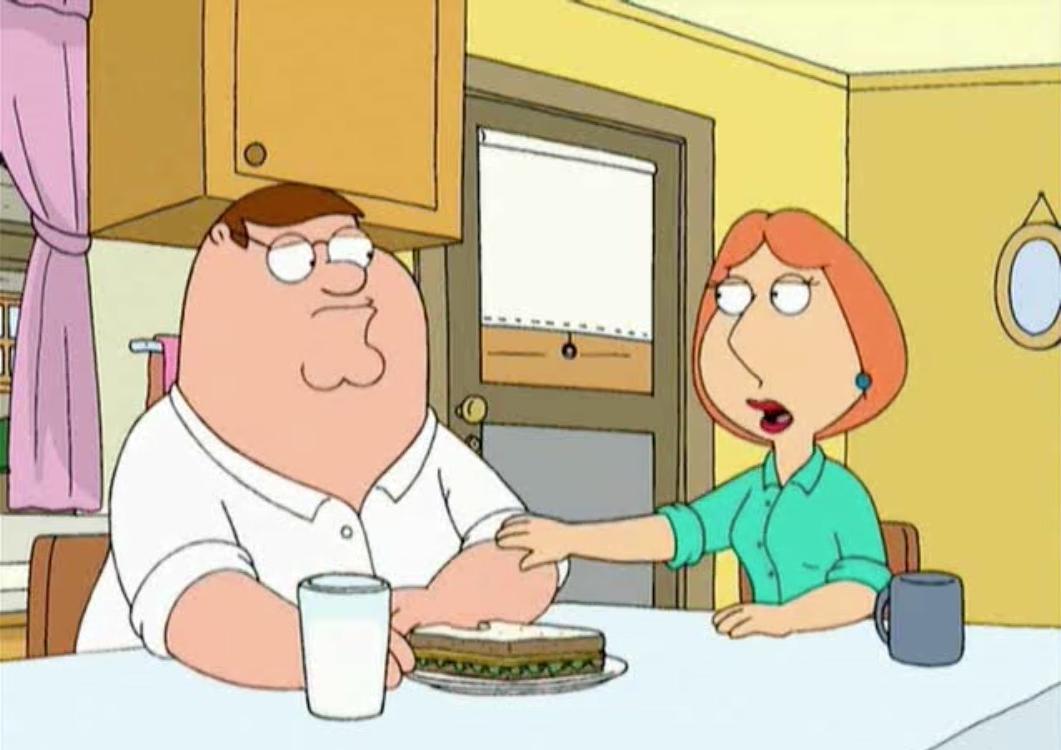}} 
						{\includegraphics[width=1.1cm,height=8mm]{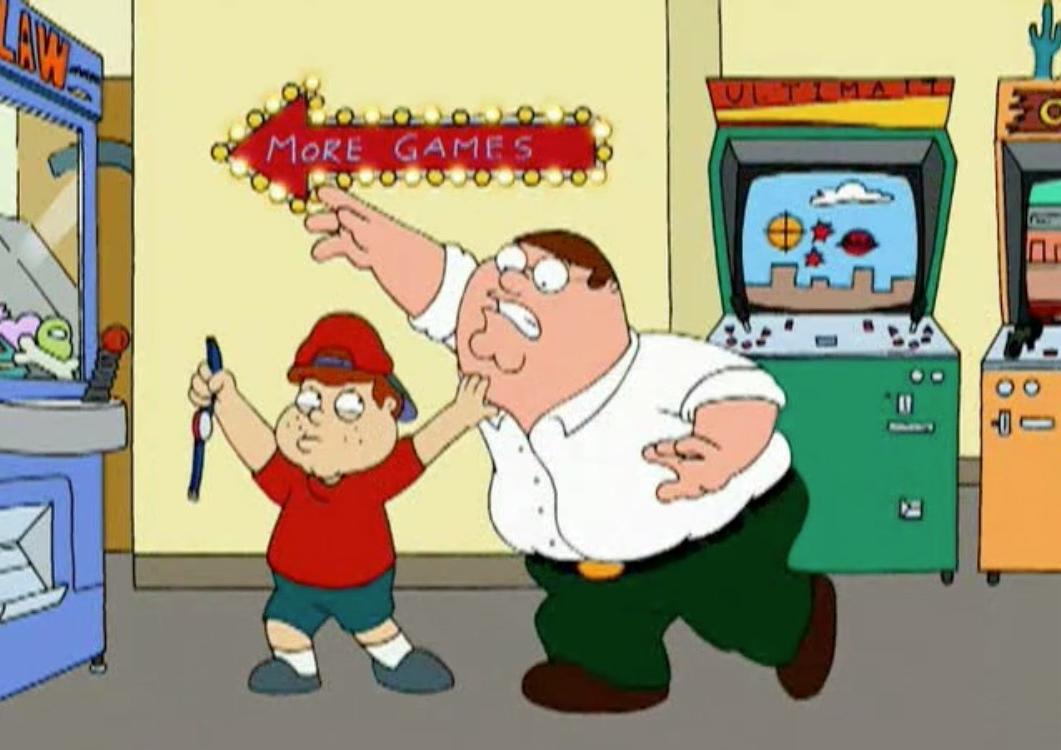}} 
						{\includegraphics[width=1.1cm,height=8mm]{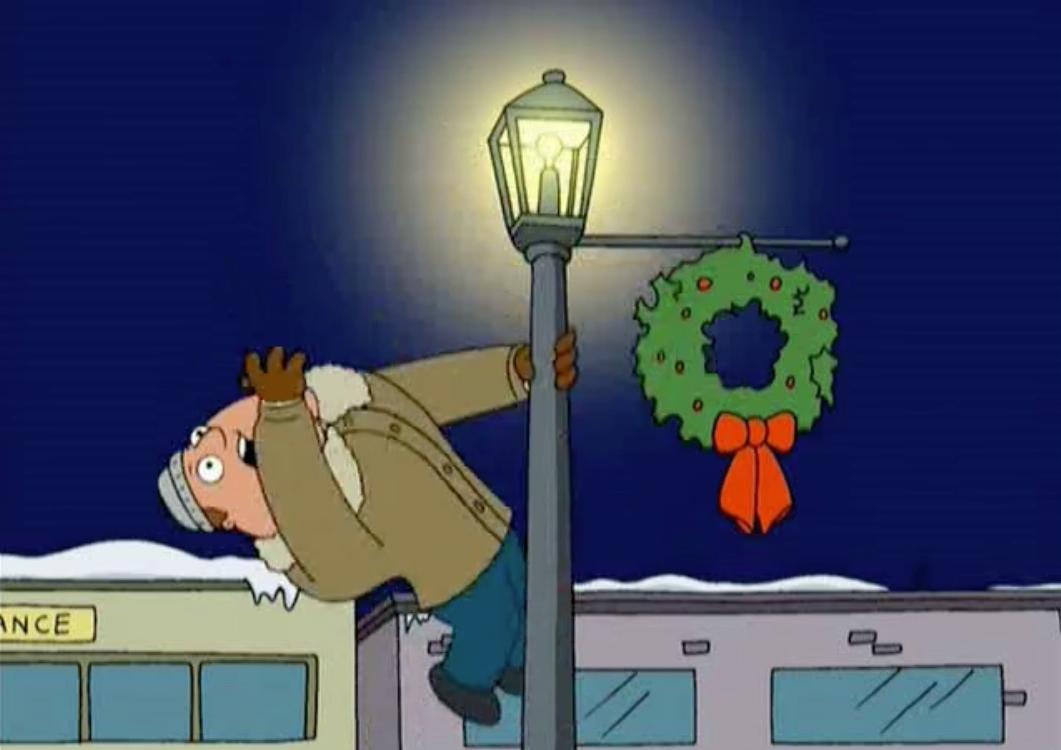}} 
						{\includegraphics[width=1.1cm,height=8mm]{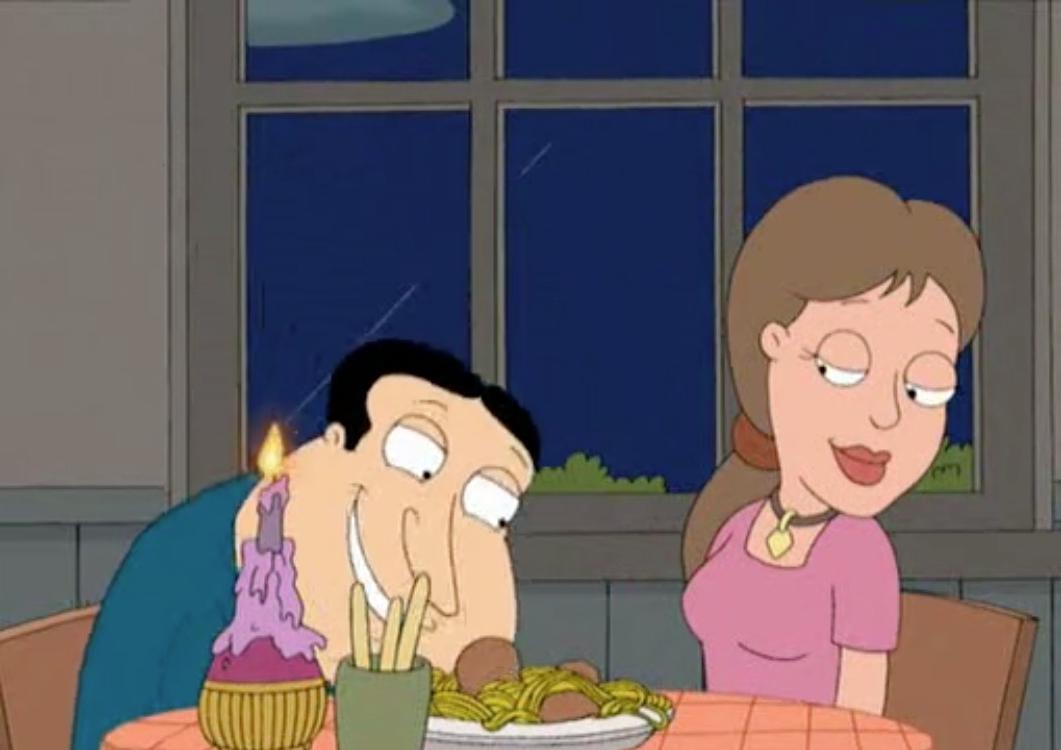}} 
						{\includegraphics[width=1.1cm,height=8mm]{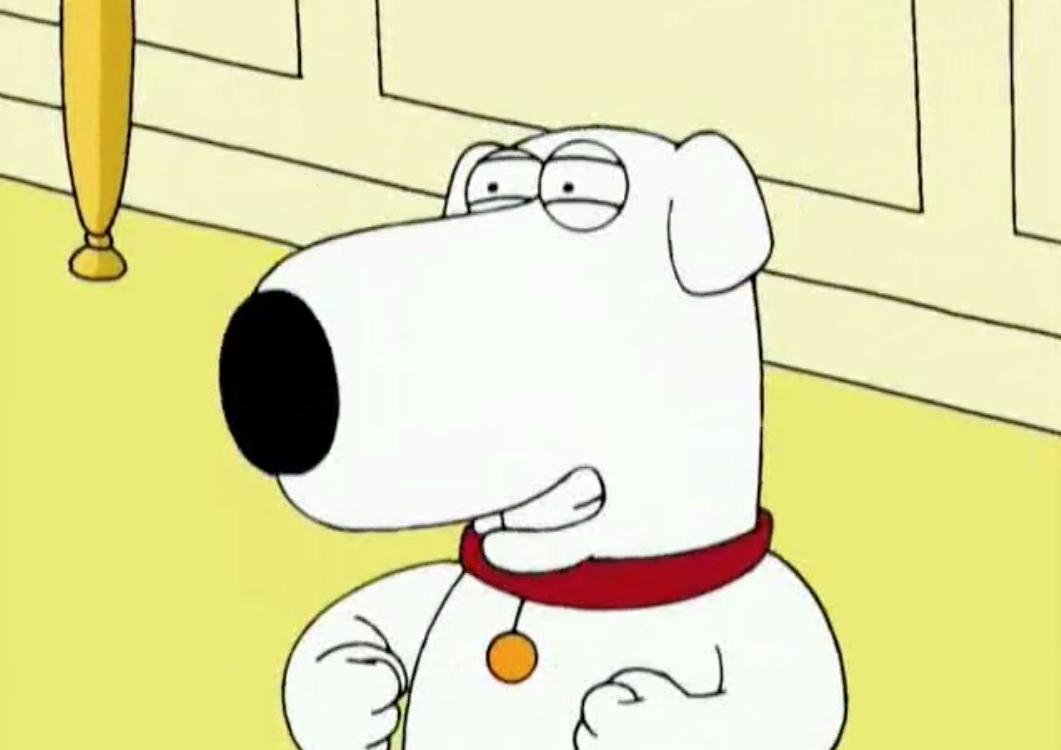}} 
						{\includegraphics[width=1.1cm,height=8mm]{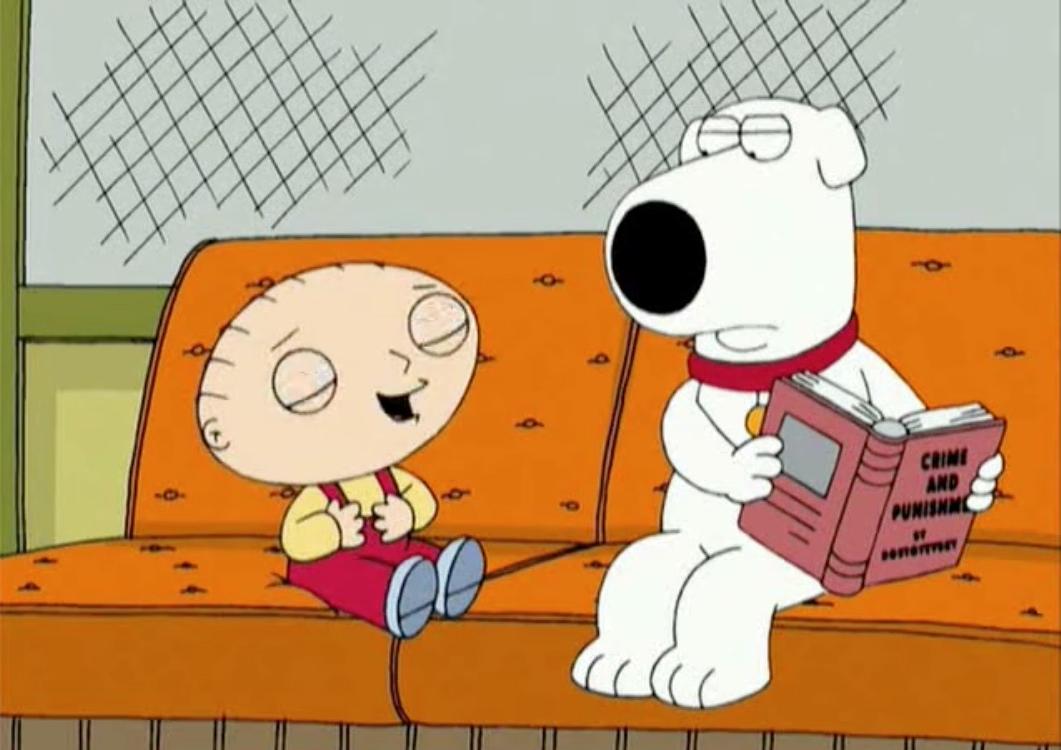}} 
						{\includegraphics[width=1.1cm,height=8mm]{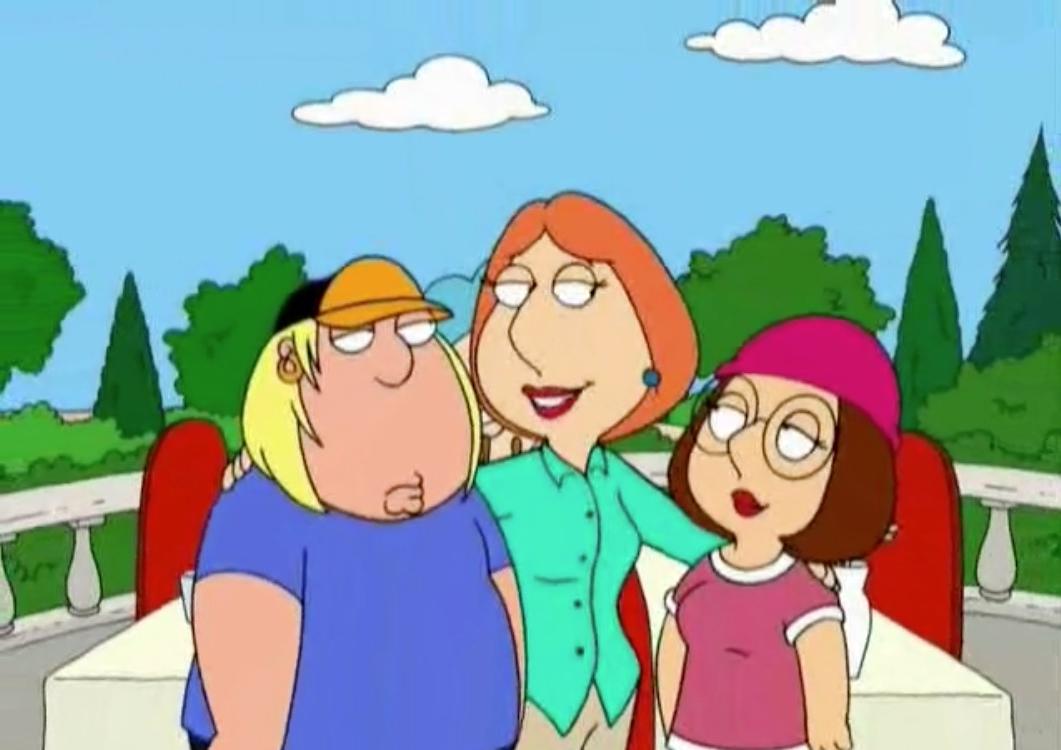}}  
						{\includegraphics[width=1.1cm,height=8mm]{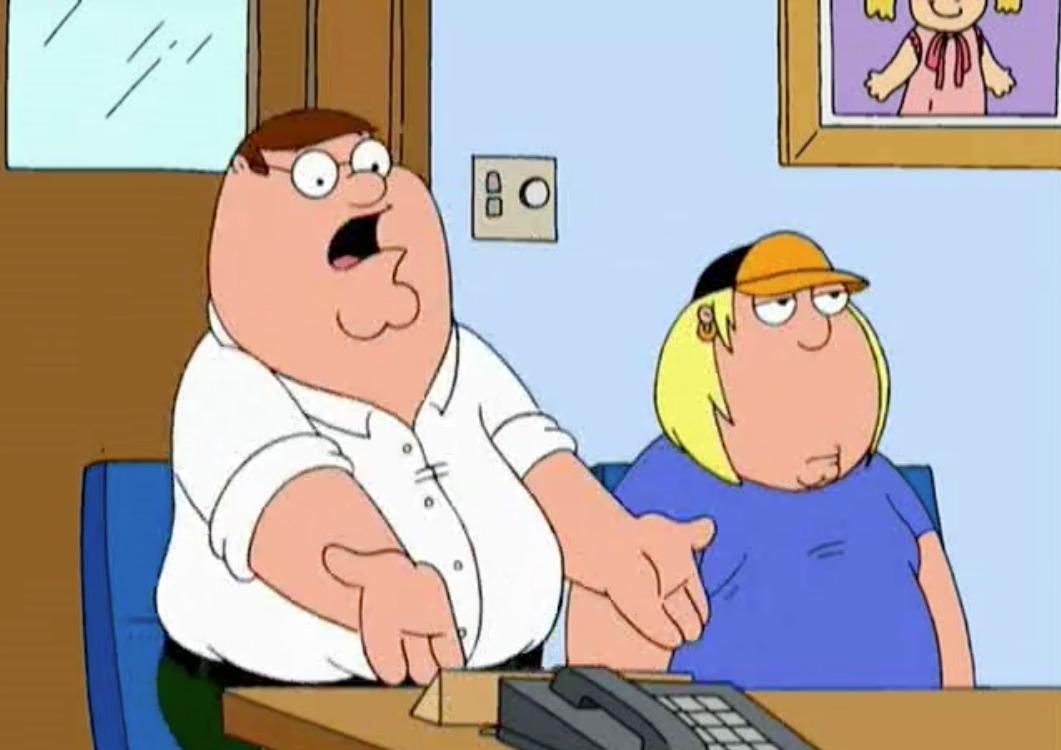}}  
					\end{flushleft}
					&
					\vspace{18pt}
					\hspace{-1pt}...\, $\rightarrow$ & 
					\vspace{0pt}
					\vspace{-6pt}
					{{{\includegraphics[height=16mm,width=2.4cm,clip,trim=0cm 2.25cm 11.3cm 0cm]{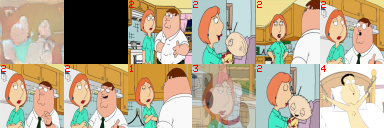}}}} \\[-8pt] \midrule
					\begin{flushleft}
						\vspace{-15pt}
						{\includegraphics[width=1.1cm,height=8mm]{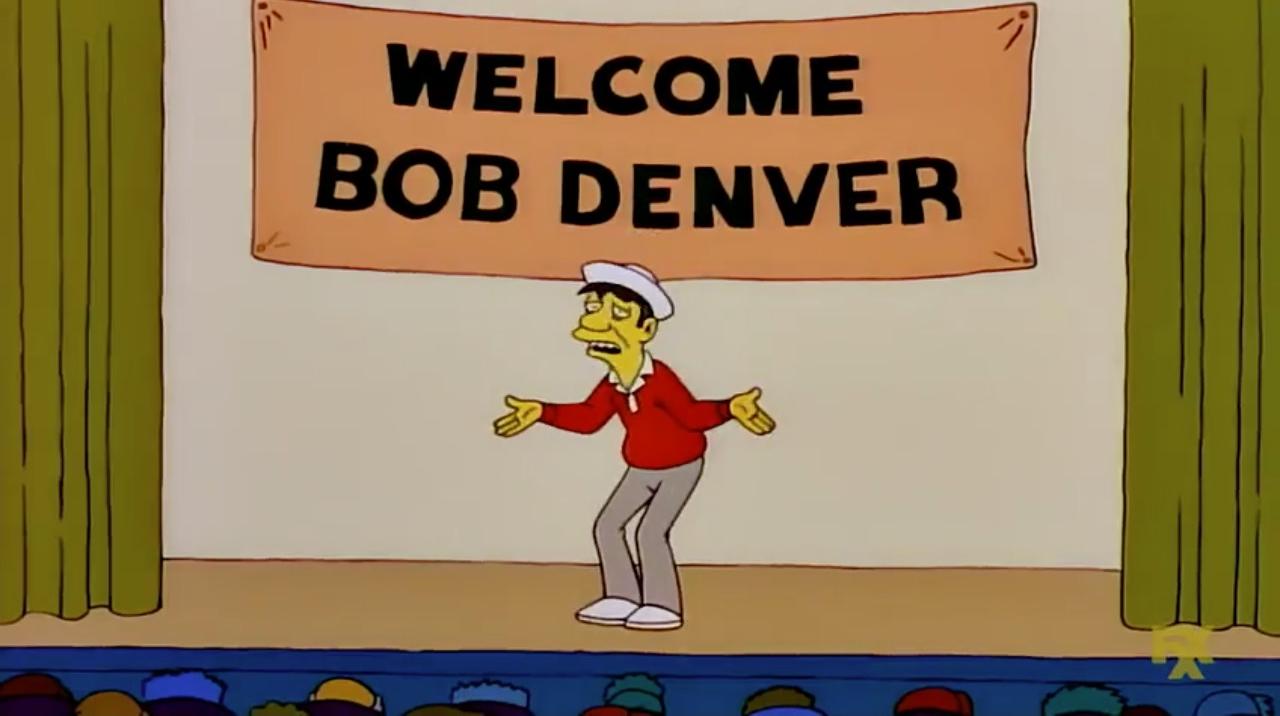}} 
						{\includegraphics[width=1.1cm,height=8mm]{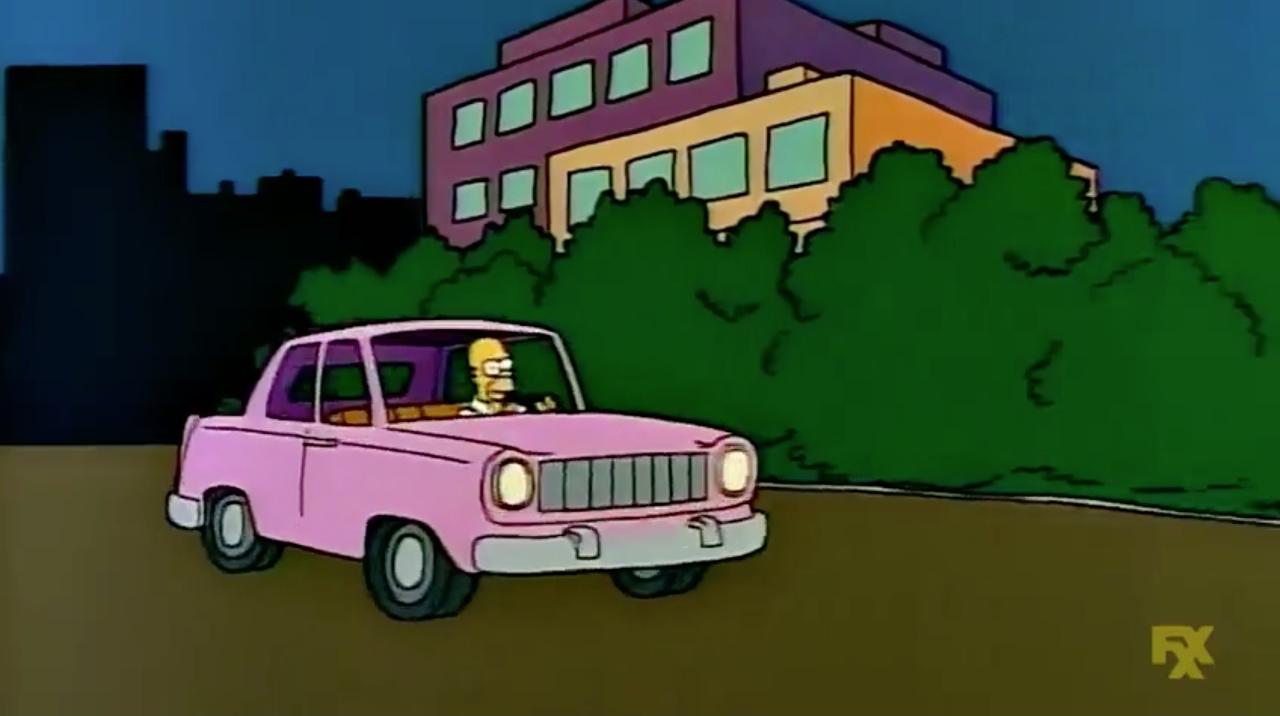}}   
						{\includegraphics[width=1.1cm,height=8mm]{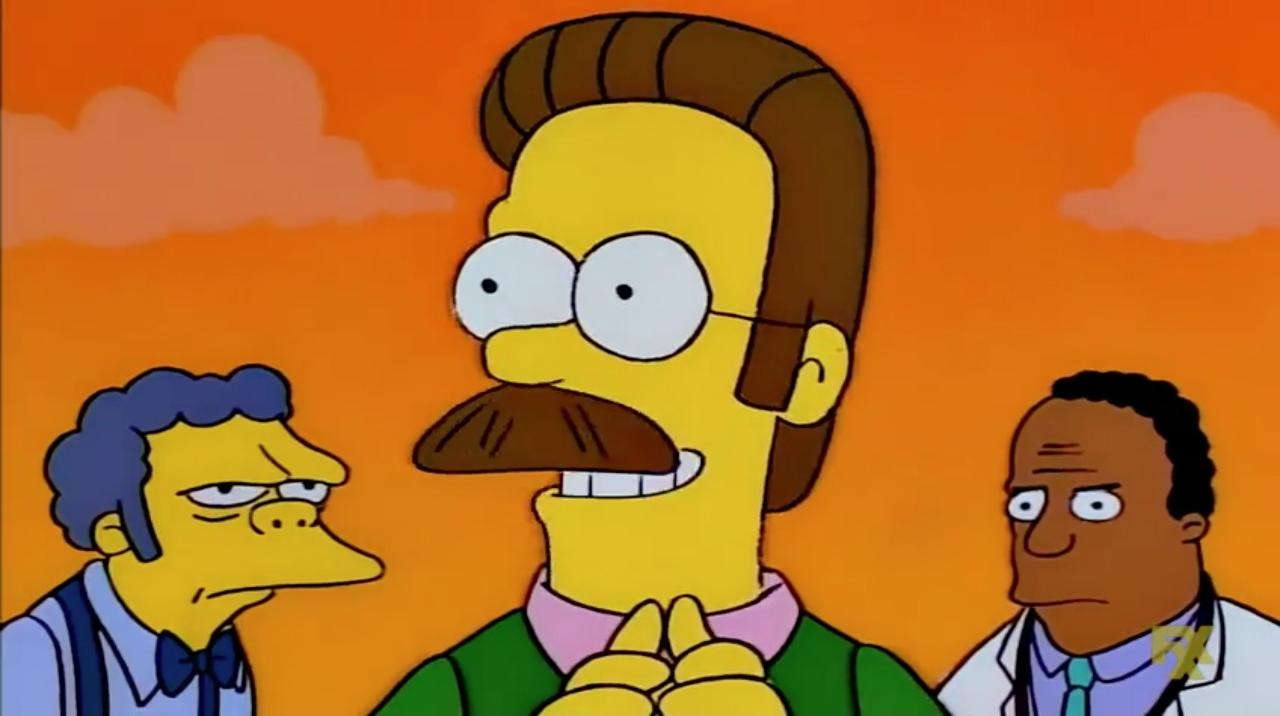}}  
						{\includegraphics[width=1.1cm,height=8mm]{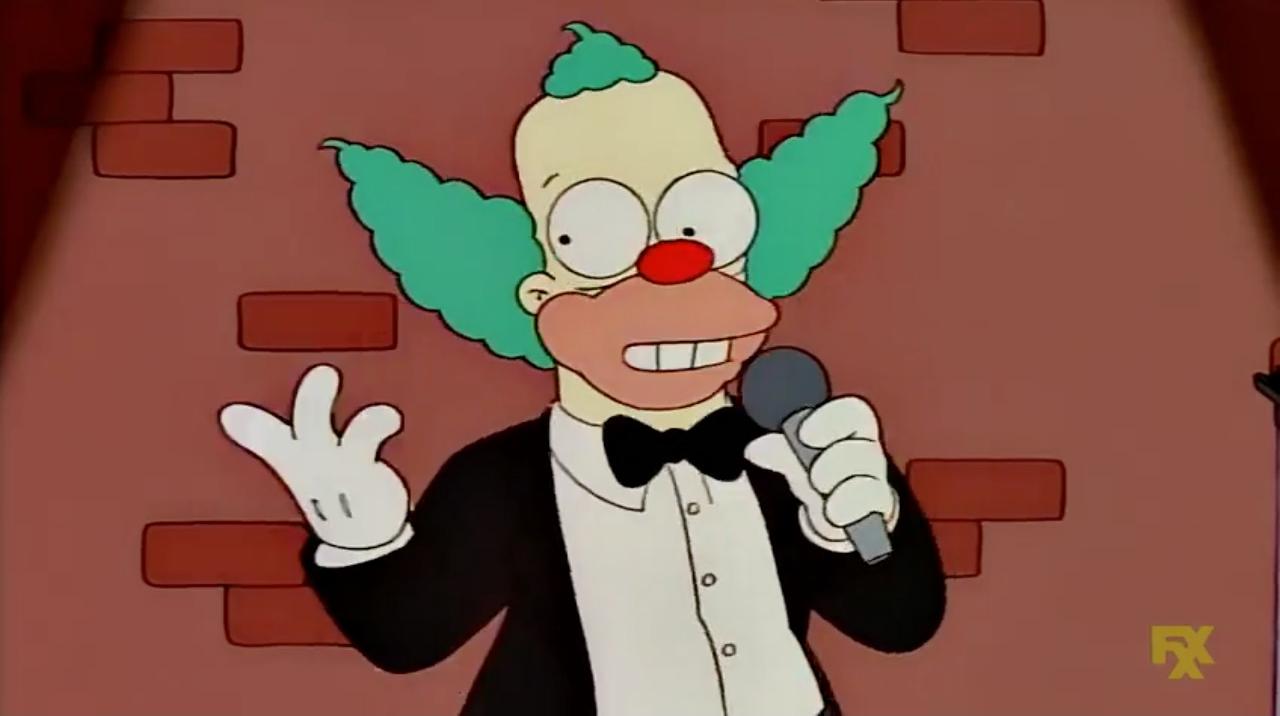}}  
						{\includegraphics[width=1.1cm,height=8mm ]{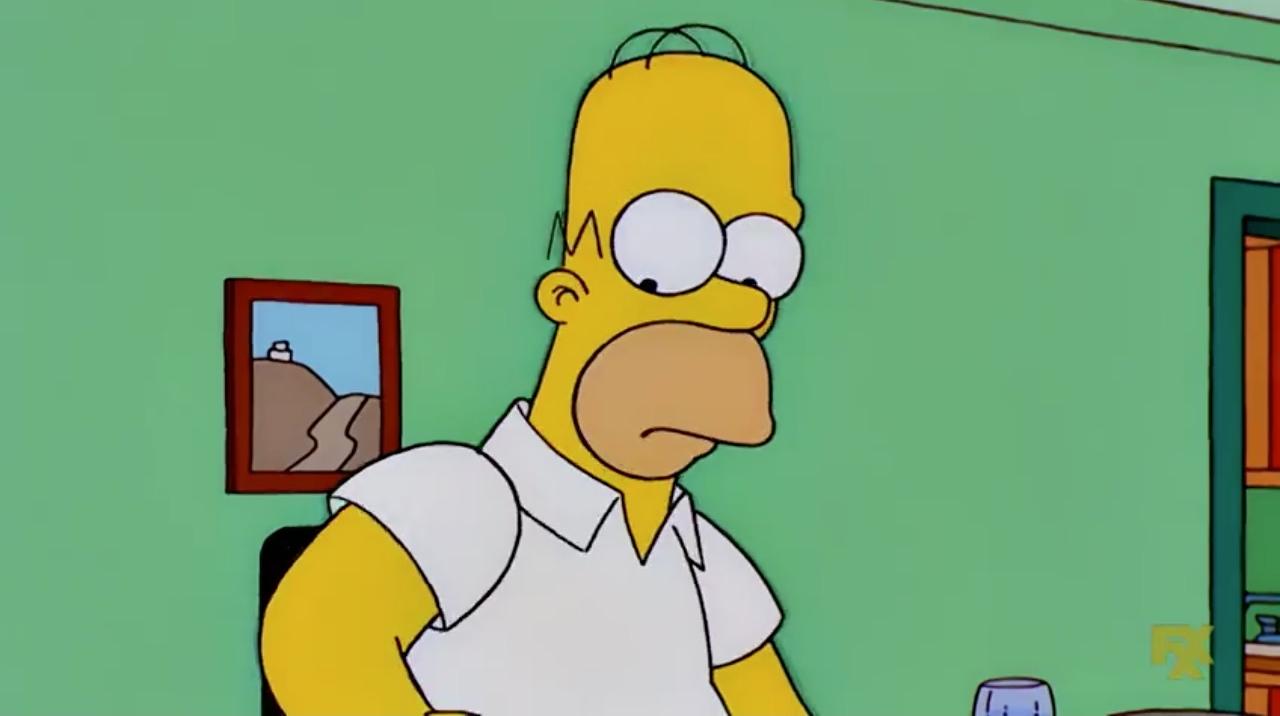}} 
						{\includegraphics[width=1.1cm,height=8mm]{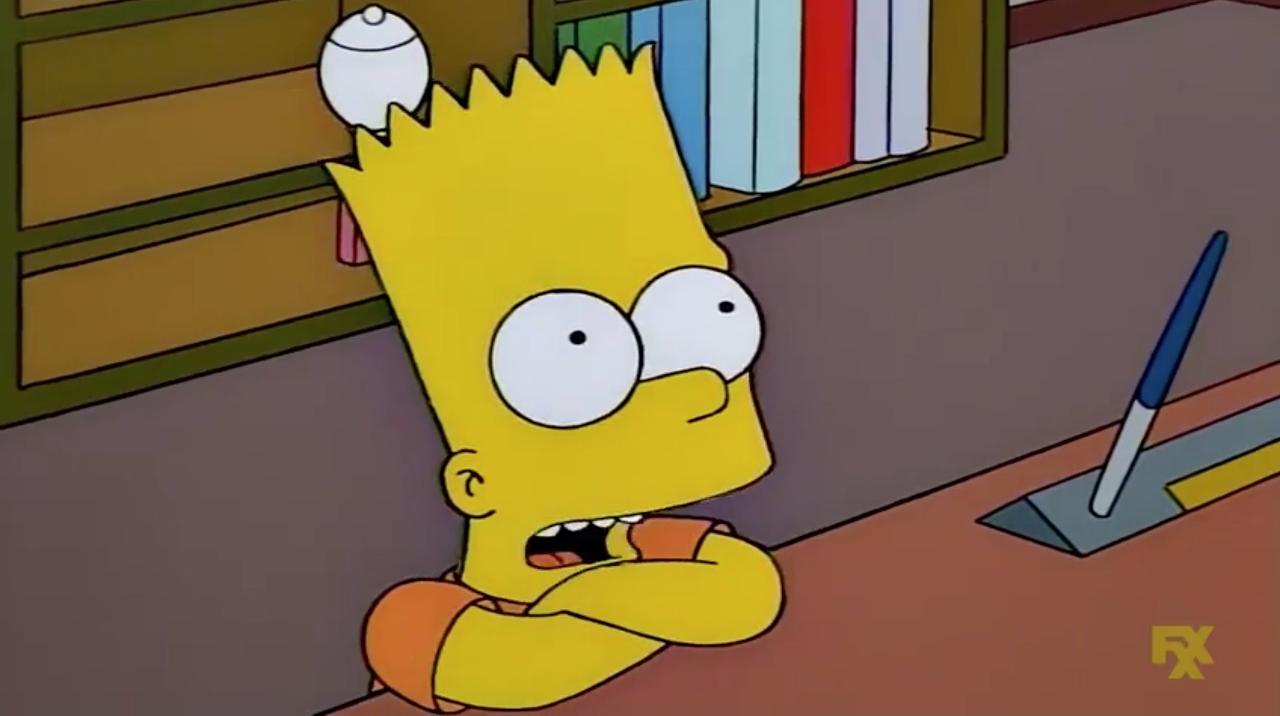}}  
						{\includegraphics[width=1.1cm,height=8mm]{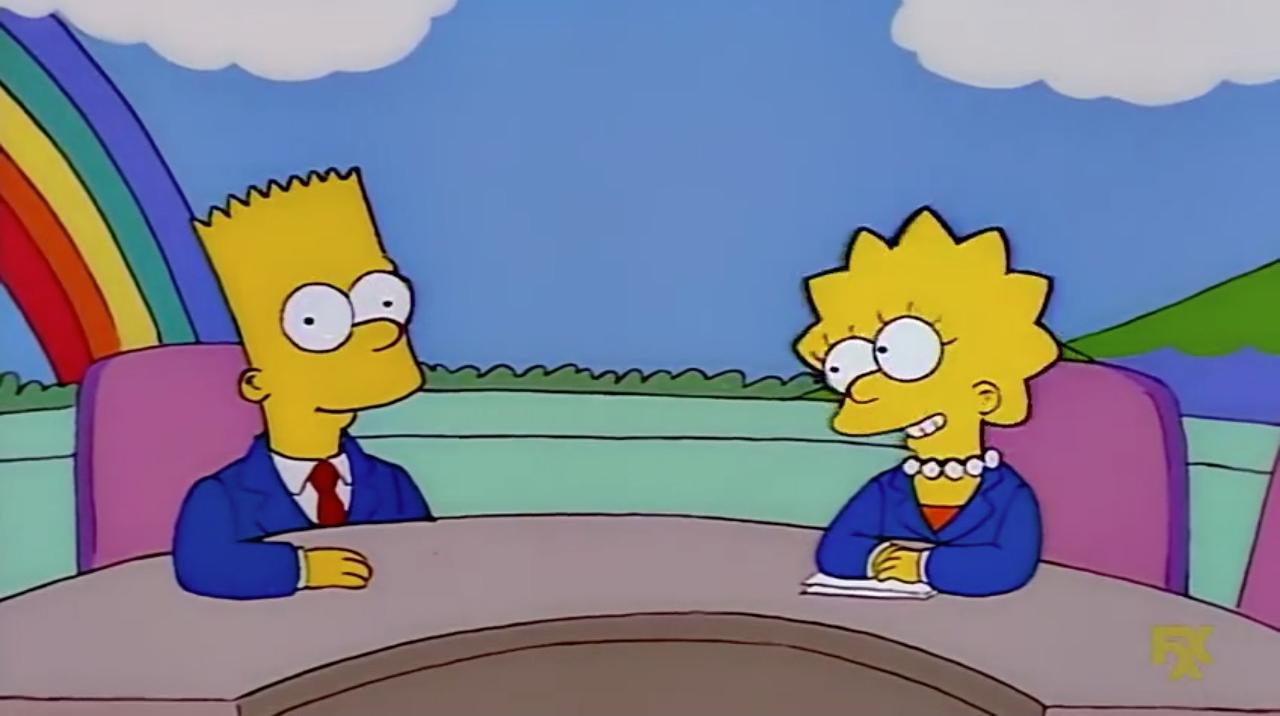}}  
						{\includegraphics[width=1.1cm,height=8mm]{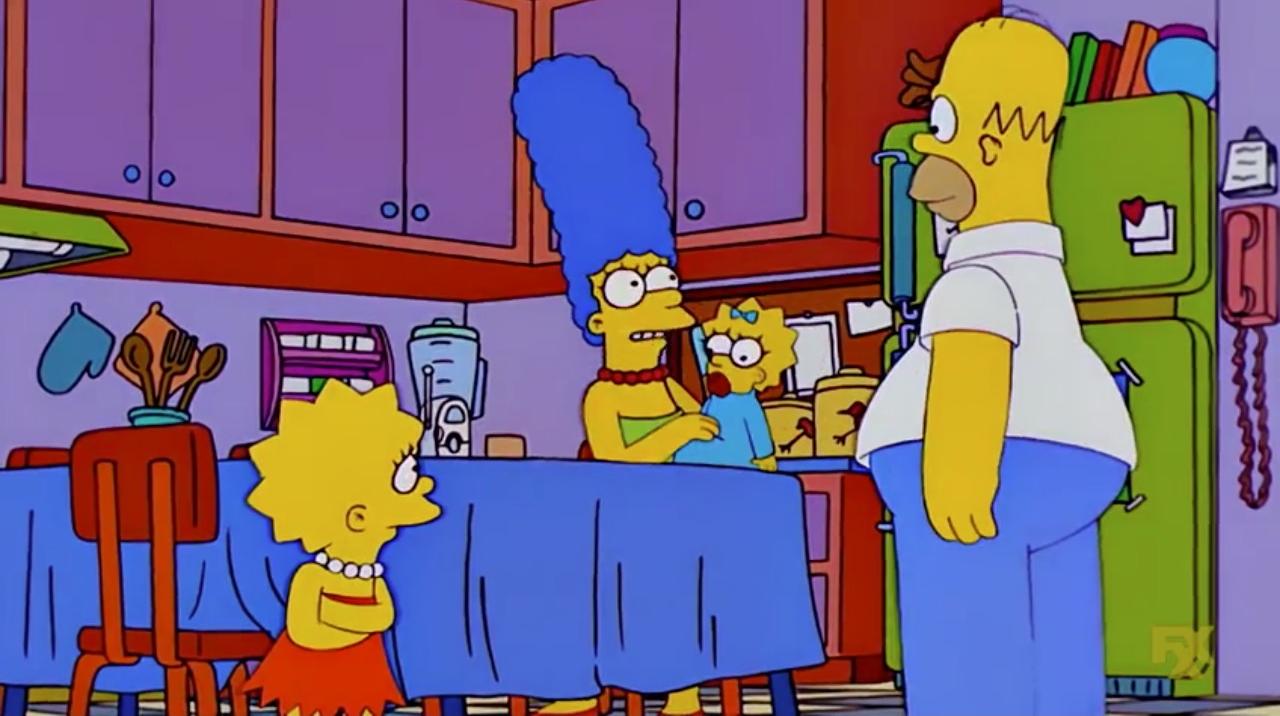}}
					\end{flushleft}
					&
					\vspace{18pt}
					...\, $\rightarrow$ & 
					\vspace{-0pt}
					\vspace{-6pt}
					{{{\includegraphics[height=16mm,width=2.4cm,clip,trim=0cm 2.25cm 11.3cm 0cm]{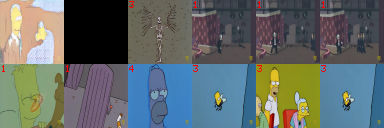}}}} 
				\end{tabular}
			}}}
			\vspace{-12pt}
			\caption{We apply Generative Adversarial Networks (GANs) to
				model the styles of two different cartoon series,
				\textit{Family Guy} (top) and \textit{The Simpsons}
				(bottom), exploring their ability to generate novel frames and find semantic relationships between the two domains.}
			\label{fig:Intro_fg}
		\end{figure}

		The fact that stories inspire one another means that the canon of film
		is full of similarities and latent connections between different works.
		These connections are sometimes obvious and
		sometimes subtle and highly debatable.  To what extent can computer
		vision find these connections automatically, based on visual features alone?
		
		As a starting point, 
		here we explore the ability of Generative Adversarial Networks (GANs) to model  the style 
		of long-running television \textit{series}. 
		GANs have shown impressive results
		in a wide range of image generation problems, including
		infilling~\cite{GANContextEncoder}, automatic
		colorization~\cite{GANColor}, image
		super-resolution~\cite{ledig2016photo}, and video frame
		prediction~\cite{videoPredict,vondrick2016generating}.  Despite this
		work, many questions remain about when GAN models work well. One
		problem is that it is difficult to evaluate the results of these
		techniques objectively, or to understand the mechanisms and failure
		modes under which they operate.  Various threads of work are underway
		to explore this, including network visualization approaches
		(e.g.~\cite{NN_visual}) and attempts to connect deep networks with
		well-understood formalisms like probabilistic graphical models
		(e.g.~\cite{PGM_deepLr}).  Another approach is to simply apply GANs to
		various novel case studies that may gave new insight into when they
		work and when they fail, as we do here.
		While predicting frames from
		individual videos has been studied~\cite{vondrick2016generating}, joint models
		of entire series may offer new insights.
		TV series usually include a set of key characters that are prominently featured in almost every
		episode, and a supporting set of characters that may appear only occasionally.
		Most series feature canonical recurring backgrounds or scenes
		(e.g.\ the coffee shop in
		\textit{Friends}), along with others that occur rarely or even just once.

		In this paper, we consider the specific case of television cartoon
		series.  We use TV cartoons because they are more 
		structured and constrained than natural images: 
		they abstract out photo-realistic details in order to
		focus on high-level semantics, which may make it easier to understand
		the model learned by a network. Nevertheless, different cartoon series
		are significantly different in the appearance of characters, 
		color schemes, background sets, artistic designs, etc. In
		particular, we consider two specific, well-known TV cartoon series:
		Simpsons and Family Guy. As our dataset, we sampled about 32,000 
		frames from four TV seasons (about 80 videos) from each series
		(Figure \ref{fig:Intro_fg}).
		
		We use these two different cartoon series to explore several
		questions.  To what extent can a network generate novel frames in the
		style of a particular series? Can a network automatically discover
		high-level semantic mappings between our two cartoon series, finding
		similar styles, scenes, and themes? If we have a frame in one series,
		can we find similar high-level scenes (e.g.\ ``people shaking hands'')
		in the other? Can training from two different series be combined to
		generate better frames from each individual series? 
		We
		find, for example, that training both domains together can generate
		better high-resolution images than either independently.
		Our work follows
		others that have considered mapping problems like image style
		transfer~\cite{gatys2016image,berkelyGANCycle} and text-to-image
		mappings~\cite{reed2016txt2img,I2T2I_ref}.
		Our three 
		contributions are: (1) proposing cartoon series as a fun, useful test domain for
		GANs, (2) building a structured but highly nontrivial mapping problem
		that reveals interesting insights about the latent space 
		produced by the GAN, (3) presenting extensive experimentation where we
		vary the training dataset composition and generation techniques in
		order to study what is captured by the underlying latent space. 
\section{Related work}

\textbf{Generative networks} learn
an unsupervised model of a 
domain such as images or videos, so that the model can generate
samples from the latent representation that it learns. For example,
Dosovitskiy \textit{et al.}~\cite{Chairsdosovitskiy2015} proposed a
deep architecture to generate images of chairs based on a latent
representation that encodes chair appearance as a function of attributes
and viewpoint.  Generative Adversarial
Networks (GANs)~\cite{goodfellow2014generative} are a particularly
prominent example. GANs model general classification problems as
finding an equilibrium point between two players, a generator and a
discriminator, where the generator attempts to produce ``confusing''
examples and the discriminator attempts to correctly classify
them. 
Radford \textit{et al.}\cite{radford2015unsupervised} combined
CNNs with GANs in their Deep
Convolutional GANs (DCGANs).
GANs have been applied
to domains including
images~\cite{mirza2014conditional,GAN_laplace},
videos~\cite{vondrick2016generating}, and even
emojies~\cite{taigman2016unsupervised}, and 
applications including  face
aging~\cite{GANFaceAge},
robotic perception~\cite{RobotocDAGan}, 
colorization~\cite{GANColor},
color correction~\cite{waterGan},
editing~\cite{GAN_editing}, and
in-painting~\cite{GANContextEncoder}.

\textbf{Conditioned Generative Adversarial Networks
	(CGANs)}~\cite{mirza2014conditional} introduce a class label in both
the generator and discriminator, allowing them to generate images with
some specified properties, such as object and part
attributes~\cite{reedDrawGAN}.  GAN-CLS~\cite{reed2016txt2img} uses
recurrent techniques to generate text encodings and DCGANs to generate
higher-resolution images given input text descriptions, while Reed
\textit{et al.}~\cite{reed2016generating} use text, segmentation masks,
and part keypoints~\cite{reed2016generating}.
Dong \textit{et
	al.}~\cite{I2T2I_ref} use an image captioning module for textual
data augmentation to enhance the performance of GAN-CLS. Perarnau
\textit{et al.}~\cite{IGANinvertible} modify the GAN architecture to
generate images conditioned on specific attributes by training an
attribute predictor, while InfoGAN~\cite{infogan} learns a more
interpretable latent representation.

\textbf{Stacked GAN architecture.} To generate higher-quality
images, coarse-to-fine approaches called
Stacked GANs have been
proposed. Denton
\textit{et al.}~\cite{GAN_laplace} describe a Laplacian pyramid GAN
that generates images hierarchically:
the first level generates a low resolution image,
which is then fed into the subsequent stage, and so on. Zhang \textit{et
	al.}~\cite{stack_gan} propose a two-stage GAN in which the first
generates a low resolution image given a text input encoding,
while the second improves its quality.
Huang \textit{et al.}~\cite{stackedGanCornel} propose a
multiple-stage GAN  in which each stage is responsible for
inverting a discriminative bottom-up mapping that is pre-trained for
classification. Wang \textit{et al.}~\cite{styleGAN} describe two
sequential GANs, one that generates surface normals and a second
that transforms the surface normals into an indoor image. Yang \textit{et
	al.}~\cite{LRGANVirgTech} describe a recursive GAN that
generates image backgrounds and foregrounds separately, and then combines them together.

\textbf{GAN Framework modifications.} The original GANs proposed by
Goodfellow \textit{et al.}~\cite{goodfellow2014generative} use a binary
cross entropy loss to train the discriminator network. 
Zhao \textit{et al.} \cite{GANenergy} employ energy-based
objectives such that the discriminator tries to assign high
energy to the generated samples, while the generator attempts to
generate samples with minimal energy. It uses an auto-encoder
to enhance the stability of the GAN. 
Metz~\textit{et al.}~\cite{unrolledGAN} define the generator objective
based on unrolled optimization of the discriminator by optimizing a
surrogate loss function. Mao \textit{et al.}~\cite{leastSQgan} address
the problem of vanishing gradients of the discriminator by proposing a
least squares loss for the discriminator instead of a sigmoid
cross-entropy. Salimans \textit{et
	al.}~\cite{salimans2016improved} propose several training strategies
to enhance the convergence of GANs, such as feature matching mini-batch
discrimination. Nowozin \textit{et al.}~\cite{nowozin2016f} consider
f-divergence measures for training generative models by regarding the
GAN as a general variational divergence estimator. Tong
\textit{et al.}~\cite{GANModeReg} address instability of the
network and the model-collapse problem by introducing two types of
regularizers: geometric metrics (e.g., the pixelwise
distance between the discriminator features and VGG features) and mode
regularizers to penalize missing modes.

\textbf{Image-to-Image GANs.} Many researchers have proposed GAN-based
image-to-image translation techniques, most of which require training data
with correspondences between images.  
Isola \textit{et al.}~\cite{isola2016image} propose conditional GANs
that map a random vector $z$ and an input image from a source domain
to an image in the target. Sangkloy \textit{et
	al.}~\cite{scribblerGAN} 
synthesize
images from rough sketches. Karacan
\textit{et al.}~\cite{layoutGAN} generate realistic outdoor scenes
based on input image layout and scene attributes. Other work
has tried to train without image-to-image correspondences in the training dataset.
For example, Taigman \textit{et al.}~\cite{taigman2016unsupervised} use
a Domain Transfer Network (DTN) with a compound loss function for mapping
between face emojis and face images.
Kim \textit{et al.}~\cite{crossDomainGAN} propose
DiscoGAN, which employs a generative model to learn a bijection 
between two domains based on loss functions that reduce the distance
between the input image and the inverse mapping of the generated
image.  Zhu \textit{et al.}~\cite{berkelyGANCycle} employ
cycle-consistent loss to train adversarial networks for image-to-image
translation.

%\newcommand{\djc}[1]{{\hl{\textbf{#1}}}}
%%---------------------------------------------------------------------------------------------------------------------------------
\section{Approach}
\label{sec:related}

Our goal is to study GANs for image generation across two different image domains --
in particular, two TV cartoon series.
While the last section gave an overview of GANs, we now focus on 
the techniques to address this particular 
task.

\begin{figure*}[th]
	{\small{
			\begin{center}
				\begin{tabular}{ccc}
					{\includegraphics[width=0.31\textwidth]{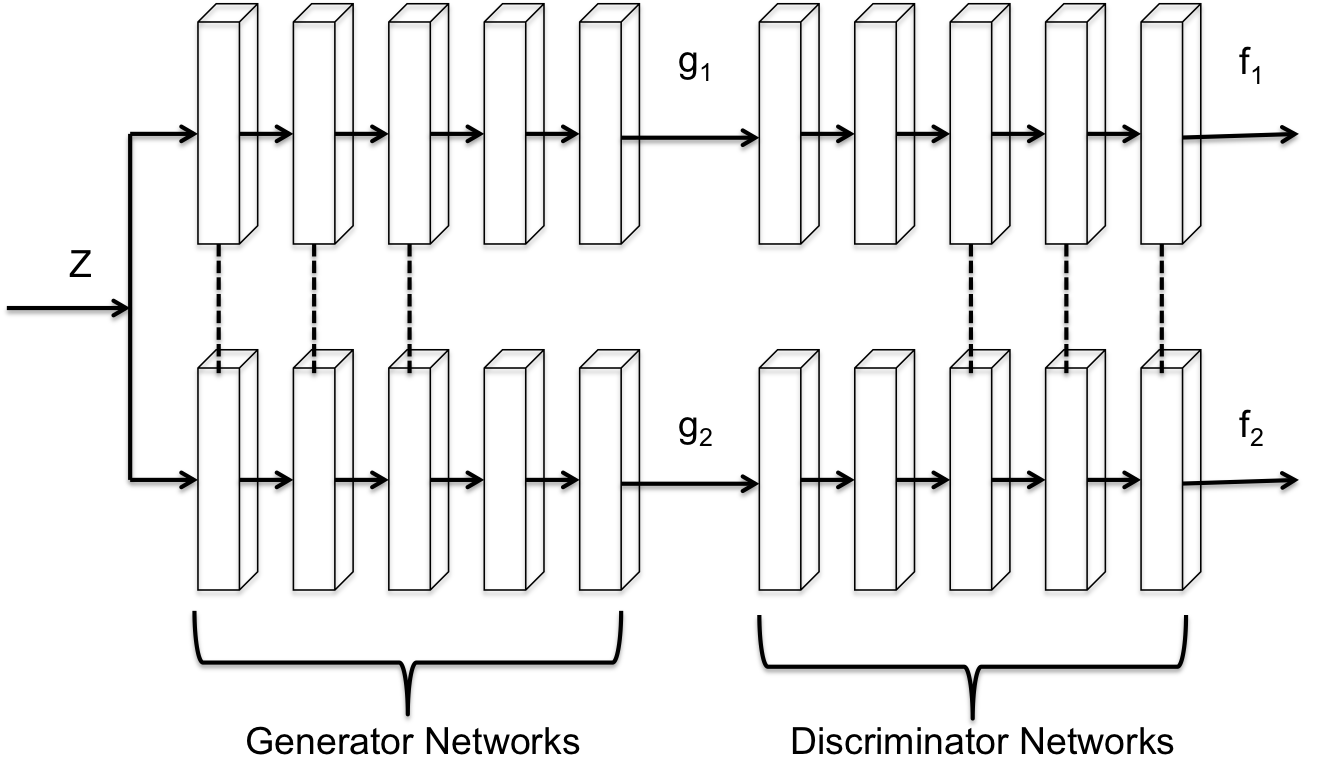}}    &
					{\includegraphics[width=0.31\textwidth]{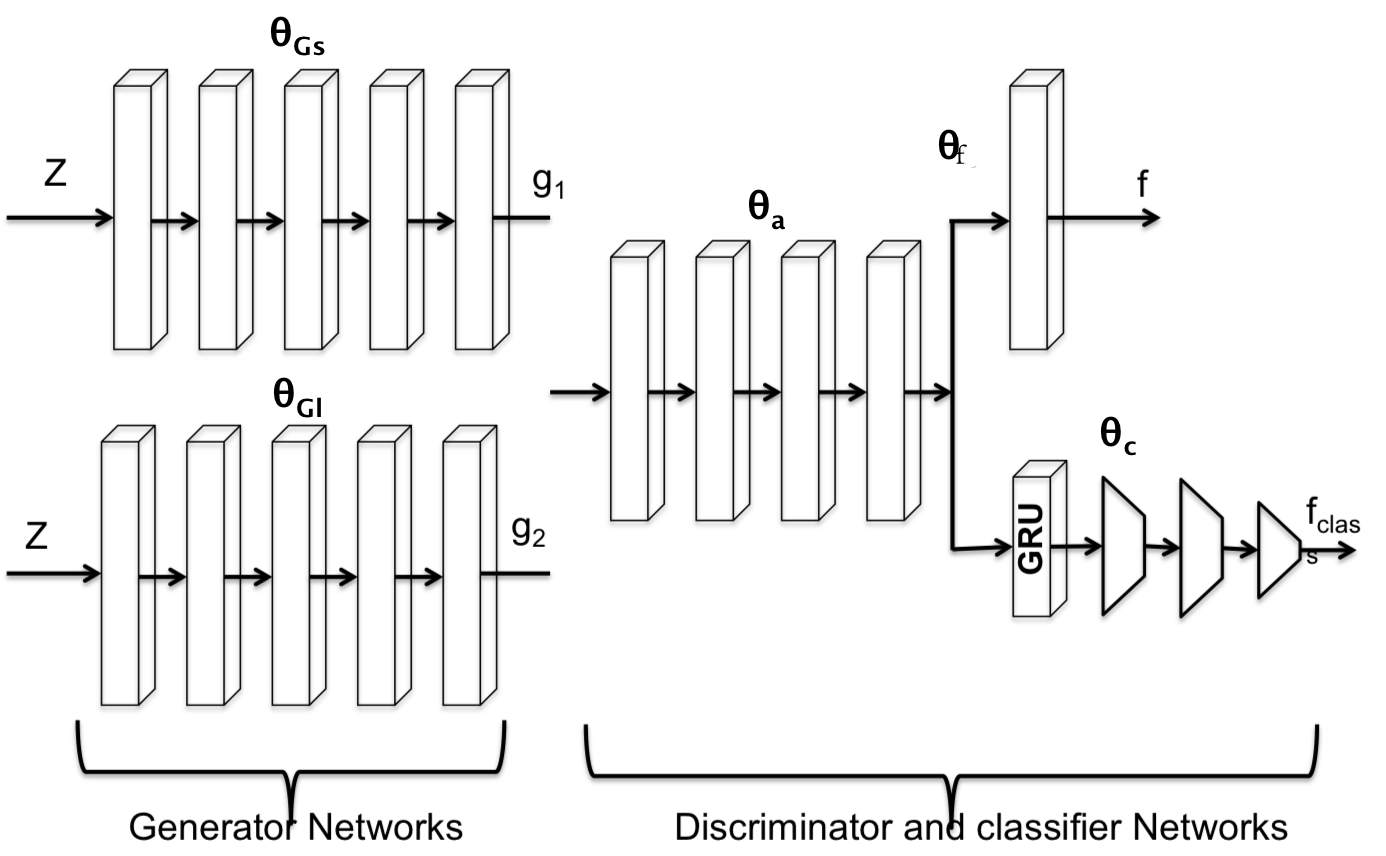}}   &
					{\includegraphics[width=0.31\textwidth]{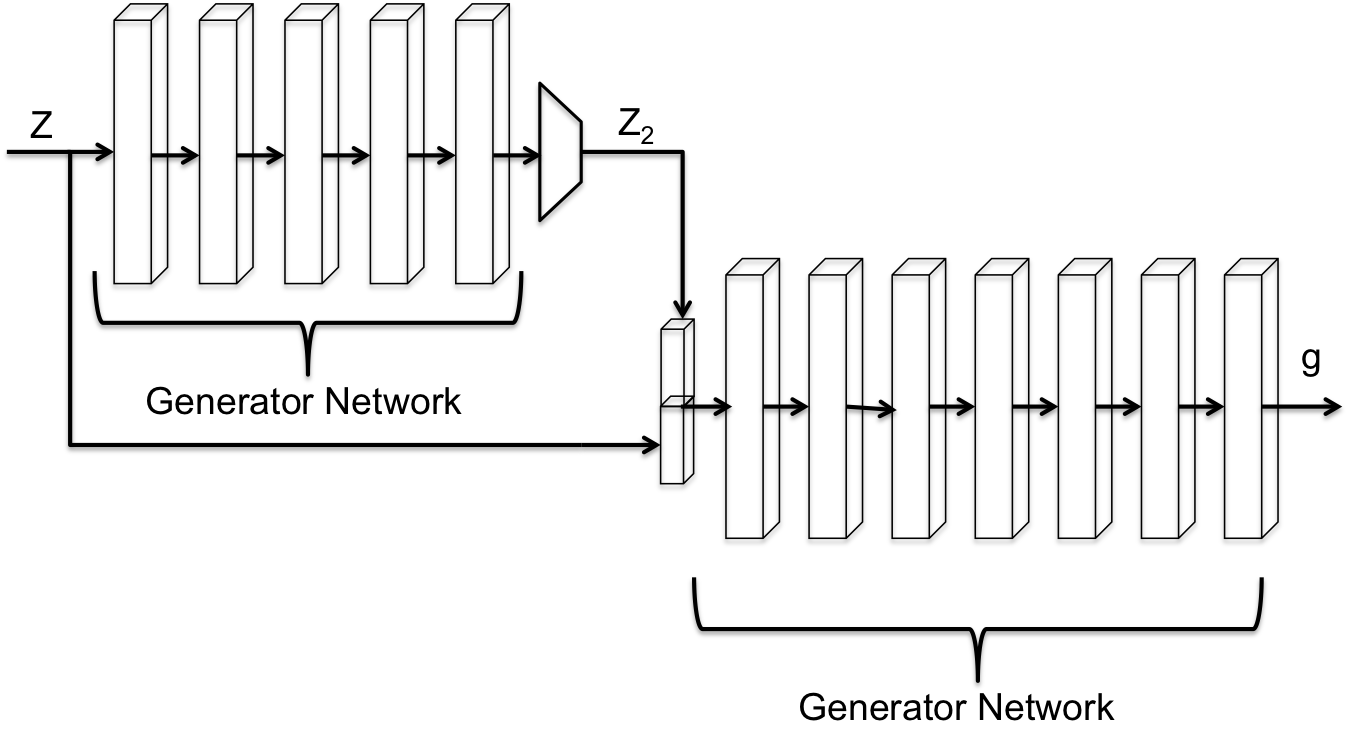}}\\
					(a) Coupled GANs & (b) GANs with domain adaptation & (c) Generator for high-resolution images
				\end{tabular}
			\end{center}
		}}
		\vspace{-10pt}
		\caption{Various network architectures that we consider.}
		\label{fig:net_all}
	\end{figure*}

	\subsection{Generative Adversarial Networks}
	
	Goodfellow~\cite{goodfellow2014generative} proposed a deep  generative model 
	as a min-max game between two agents: (1)
	a generative network $g(z;\theta_g)$ that models a
	probability distribution $p_g$ of generated image samples with input
	$z$ that has distribution $z \sim p_z$, and (2) a
	discriminator network $f(x;\theta_f)$ which estimates the probability
	that the input sample is drawn from the real data distribution $x \sim
	p_x$ or from the generated distribution $g(z; \theta_g)$. Ideally $f(x;\theta_f) = 1$ if
	$x \sim p_x$ and $f(g(z; \theta_g);\theta_f) = 0$ if $z \sim p_z$. The networks
	are trained by optimizing,
	$$ \min_{\theta_g}\max_{\theta_f} L(\theta_f,\theta_g), $$
	
	\begin{align*}
	\begin{split}
	L(\theta_f,\theta_g)  = & E_{x\sim p_{x}}[\log(f(x;\theta_f))] + \\
	& E_{z\sim p_{z}}[\log(1-f(g(z;\theta_g); \theta_f))].
	\end{split}
	\end{align*}
	The optimization problem is solved by alternating between gradient ascent steps
	for the discriminator,
	$$            \theta_{f}^{t+1} = \theta_{f}^{t} + \lambda^t \bigtriangledown_{\theta_{f}} L(\theta_f^t,\theta_g^t),  $$
	and descent steps on the generator,
	$$            \theta_{g}^{t+1} = \theta_{g}^{t} -\lambda^t \bigtriangledown_{\theta_{g}} L(\theta_f^{t+1},\theta_g^t). $$
	
	%------------------------------------------------------------
	\subsection{DCGAN}
	Deep Convolutional GANs (DCGANs)
	\cite{radford2015unsupervised} include architectural constraints
	for deep unsupervised network training,  many of which we
	follow here. Any deep architecture we employ
	consists of a set of modules (represented by rectangular blocks in
	network architectures). Each module in the generator consists of
	fractional-strided convolutions, batch normalization and ReLU
	activation for all layers except for the output, which uses hyperbolic tangent. In
	the discriminator, each block consists of strided convolutions, batch
	normalization and LeakyReLU activations.
	
	%------------------------------------------------------------
	\subsection{CO-GAN}
	
	Coupled GANs (Co-GANs)~\cite{liu2016coupled} allow for a network to model
	multiple image domains. 
	A CO-GAN
	consists of two (or more) GANs,
	where $(g_s,f_s)$ and $(g_l,f_l)$
	represent the generative and discriminative network for the first and
	second domains, respectively (Figure~\ref{fig:net_all}(a)).
	Since the first layers of the discriminator encode
	low-level features while the later layers encode high-level
	features, 
	%\djc{is it OK to use WE ??} %djc: Yes!
	we force the domains to share 
	semantic information by tying the weights of their final discriminator layers together.
	Since the flow of information in
	generative networks is opposite (initial layers represent high-level concepts),
	we 
	tie together the early layers of the generators.
	In learning, Co-GANs
	solve a constrained minimax game similar to that of GANs,
	$$ \max_{\theta_{g_s},\theta_{g_l}}\min_{\theta_{f_s},\theta_{f_l}} L(\theta_{f_s},\theta_{g_s},\theta_{f_l},\theta_{g_l}), $$.
	We will use CO-GANs to explore semantic connections between Family Guy and The Simpsons and compare the results with other methods described in the following section.

	\subsection{Adversarial domain adaptation  } 
	
	In domain adaptation, we have an input space $X$ (images) and
	output space $Y$ (class labels). The objective is to train the
	model using source domain distribution $S(x,y)$ so as to maximize the
	performance on a target domain distribution $\tau(x,y)$. Both
	distributions are defined on $X \times Y$ where $S$ is ``shifted'' from
	$\tau$ by some domain offset.
	
	The adversarial domain adaptation model~\cite{DA_unsup} decomposes the
	input image feature representation into three parts: $G_{ft}$, which
	extracts feature representations $ft = G_{ft}(x; \theta_{ft})$,
	$G_y(ft;\theta_y)$, which maps a feature vector $ft$ to label $y$, and
	$G_d(ft;\theta_d)$, which maps a feature vector $ft$ to 
	either 0 or 1 according to whether it belongs to the source or target distribution, respectively.
	The objective is to obtain a domain invariant
	feature representation $f$ that maximizes the loss of the domain
	classifier while making the parameters $\theta_d$ minimize the loss of
	the domain classifier. 
	%\djc{That last sentence is contradictory, I
	%  think -- one of those ``domain classifier'' should be changed,
	%  right? -- no that is why GRU add -ve sign to the back-propagated gradient, so when the loss of domain classifier propagated theta d is optimized to minimize this loss till the gru layer that inverts the gradient propagated so the proceeding layers that produce f maximize the domain classifier loss, that is why the original paper called this method adversarial domain adaptation } 
	This is accomplished by a gradient reversal unit (GRU),
	which acts as identity during forward propagation and value negation
	during backward propagation.
	
	\subsection{ Domain adaptation for generative models }
	
	Consider two distributions, $X_s \sim \tau(x_s)$ and $X_l \sim
	\tau(x_l)$, corresponding to two different input domains, and a set of
	samples from each domain, $\{x_{1s},x_{2s}, \hdots x_{sN} \}$ and $\{
	x_{1l}, x_{2l}, \hdots x_{lN} \}$. 
	Each sample has a label $y \in \{0,1\}$ corresponding to whether
	it is a fake or real image, respectively, and a label $d \in \{0,1\}$
	indicating whether it is from the first or second domain.
	%Consider our fake real labels for
	%discriminator in the GAN model for both domains $y \in \{0,1\}$ that
	%represents fake and real images respectively. Also $d \in \{0,1\}$ to
	%represent each domain (Simpson and Family Guy respectively). 
	The
	parameters for the model are $\theta_{Gs}$, $\theta_{Gl}$,
	$\theta_{f}$, $\theta_c$, $\theta_a$ and the network architecture is shown
	in Figure~\ref{fig:net_all}(b).
	Training involves optimizing an energy function,
	\begin{align*}
	\begin{split}
	E(\theta_{Gs},\theta_{Gl},\theta_a, \theta_{f},\theta_c) &= L_1(\theta_{Gs},\theta_{Gl},\theta_a, \theta_{f})   + L_2(\theta_c, \theta_a),
	\end{split}
	\end{align*}
	where $L_1$ is the cross entropy loss,
	\begin{align*}
	\begin{split}
	L_1(\theta_{Gs},&\theta_{Gl},\theta_a, \theta_{f}) =  \\
	&    E_{x_s \sim p_{x_s}}[\log(f(x_s;\theta_{af}))]  + \\
	&    E_{x_s \sim p_{x_l}}[\log(f(x_l;\theta_{af}))]+ \\
	&    E_{z_s \sim p_{z_s}}[\log(1-f(g_s(z_s;\theta_{g_s});\theta_{af}))]+ \\
	&    E_{z_l \sim p_{z_l}}[\log(1-f(g_l(z_l;\theta_{g_l});\theta_{af}))],
	\end{split}
	\end{align*}
	where $\theta_{af} = \{\theta_a, \theta_f\}$ and $L_2$ is a binary layer log max loss for the domain classifier,  
	$$ L_2(\theta) = -\frac{1}{N} \sum_{i=1}^{N}{ \sum_{j=1}^{k}{ 1\{y^{(i)}=j\} \log\left( \frac{e^{\theta^T_j x^{(i)}  } }{ \sum_{l=1}^{k}{ e^{\theta^T_l x^{(i)}  } } }  \right),  } } 
	$$
	with $ \theta = \{ \theta_c , \theta_a\}.$
	Algorithm~\ref{alg:alg_steps} shows the steps involved in this optimization. In the experimental results, we will examine the results of applying that technique on finding high level semantic alignments between the two domains of Simpsons and Family Guy.
	%The training steps are described in algorithm \ref{alg:alg_steps}. This algorithm describes the sequence of steps to find a local optimum for the loss function $E(\theta_{Gs},\theta_{Gl},\theta_a, \theta_{D},\theta_c)$. 
	
	\begin{algorithm}
		\caption{Training with domain adaptation}
		\label{alg:alg_steps}
		\begin{algorithmic}[1]
			{\scriptsize{
					\State Given: Minibatch size $b$, learning rate $\lambda$, iteration count $n$,
					randomly-initialized $\theta_{G_s}$, $\theta_{G_l}$, $\theta_f$, $\theta_c$, $\theta_a$.
					\For{$i = 1$ to ${n}$}
					\State Update discriminator parameters $ \{\theta_a, \theta_{f}\} $, where real images of both domains have label 1 and fake (generated) images have label $0$.
					\vspace{-8pt}
					\begin{equation}
					\{\theta_a, \theta_{f}\} \longleftarrow  \{\theta_a, \theta_{f}\} + \lambda  \frac{\partial L_1(\theta_{Gs},\theta_{Gl},\theta_a, \theta_{f})  }{\partial \theta_{ \{\theta_a, \theta_{f}\}  } } 
					\end{equation}
					\vspace{-8pt}
					\State Update generative models for the two domains independently, 
					by propagating the cross-entropy from the discriminative network, considering images generated from the generator as real,
					\vspace{-8pt}
					\begin{equation}
					\theta_{Gs} \longleftarrow  \theta_{Gs} - \lambda  \frac{\partial L_1(\theta_{Gs},\theta_{Gl},\theta_a, \theta_{f})  }{\partial \theta_{ \theta_{Gs}  } } 
					\end{equation}
					%        
					%        \State Similarly for the generative model for the family Guy domain
					\vspace{-8pt}
					\begin{equation}
					\theta_{Gl} \longleftarrow  \theta_{Gl} - \lambda  \frac{\partial L_1(\theta_{Gs},\theta_{Gl},\theta_a, \theta_{f})  }{\partial \theta_{ \theta_{Gl}  } } 
					\end{equation}
					\vspace{-8pt}
					\State Update classifier parameters ${\theta_a, \theta_c}$ with real samples from both domains, where class labels indicate domain, and propagate error through the classifier and shared discriminator parameters,
					\vspace{-8pt}
					\begin{equation}
					\{\theta_a, \theta_{c}\} \longleftarrow  \{\theta_a, \theta_{c}\} - \lambda  \frac{\partial L_2(\theta_a, \theta_{c})  }{\partial \theta_{ \{ \theta_a, \theta_{c} \}  } } |_{real}
					\end{equation}
					\vspace{-8pt}
					\State Repeat with fake generated images, 
					\vspace{-8pt}
					\begin{equation}
					\{\theta_a, \theta_{c}\} \longleftarrow  \{\theta_a, \theta_{c}\} - \lambda  \frac{\partial L_2(\theta_a, \theta_{c})  }{\partial \theta_{ \{ \theta_a, \theta_{c} \}  } } |_{fake}
					\end{equation}
					\vspace{-8pt}
					\EndFor
				}}
			\end{algorithmic}
		\end{algorithm}
		%\end{figure}

		%\djc{we should refer to Algorithm 1 somewhere in the text. Is any part of this algorithm novel?}
		
		%\djc{Also I'm not sure what this next part is trying to say. We haven't defined any algorithm yet so I'm not sure what variations we're talking about. For me it is just some set of experiments, to some how find the relevance of each part of the model for the performance of the system, is any steps are crucial or not, it happens that not necessarily, that removing any step could affect the performance dramatically. Also it is necessarily that every part  make the model learn a new mode of the mapping}
		%\djc{ A side note : is it possible that  the same mapping happens by simple features space searching without any joint training between domain to create this feature space, I really should have tested this hypothesis. Another side note : I really didn't see that work as a way to provide a good answer to the question or build a very good model to do the mapping, I just view this work as a simple trials or experiments to address the problem and explore the challenges to the problems, with another more unexplored challenge is adding subtitles as description to the image frames, is it going to make the mapping better or harder }

\newcommand{\xhdr}[1]{{\textbf{\textit{#1}}}}

\section{Experimental Results}

\newcommand{\fourbythree}[1]{\includegraphics[width=0.23\textwidth,trim=0cm 2.25cm 6.75cm 0cm,clip]{#1}}
\newcommand{\fourbythreehi}[1]{\includegraphics[width=0.23\textwidth,trim=0cm 4.5cm 13.5cm 0cm,clip]{#1}}
\newcommand{\fourbytwo}[1]{\includegraphics[width=0.23\textwidth,trim=0cm 2.25cm 9cm 0cm,clip]{#1}}
\newcommand{\fourbytwohi}[1]{\includegraphics[width=0.23\textwidth,trim=0cm 4.5cm 18cm 0cm,clip]{#1}}
\newcommand{\fourbytwovhi}[1]{\includegraphics[width=0.23\textwidth,trim=0cm 9cm 36cm 0cm,clip]{#1}}
\newcommand{\threebythree}[1]{\includegraphics[width=0.23\textwidth,trim=0cm 4.5cm 6.75cm 0cm,clip]{#1}}
\newcommand{\threebythreehi}[1]{\includegraphics[width=0.23\textwidth,trim=0cm 9cm 13.5cm 0cm,clip]{#1}}
\newcommand{\fullfourbytwo}[1]{\includegraphics[width=0.4\textwidth,trim=0cm 2.25cm 9cm 0cm,clip]{#1}}
\newcommand{\fullfourbytwohi}[1]{\includegraphics[width=0.4\textwidth,trim=0cm 4.5cm 18cm 0cm,clip]{#1}}
\newcommand{\fulltwobyfour}[1]{\includegraphics[width=0.4\textwidth,trim=0cm 6.75cm 4.5cm 0cm,clip]{#1}}
\begin{figure*}[t]
	{\small{
			\begin{center}
				\begin{tabular}{p{12pt}@{}cc|cc} 
					& \textsf{Simpsons} &\multicolumn{2}{c}{}& \textsf{Family Guy} \\ \toprule
					\begin{sideways}{~\hspace{1in}\textsf{$64 \times 64$}}\end{sideways} &
					\includegraphics[width=0.4\textwidth]{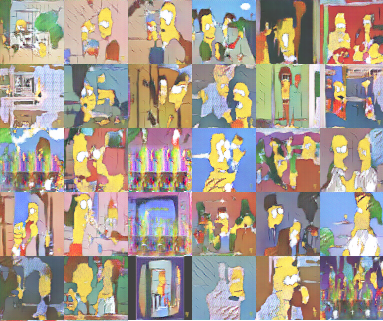} &&&
					\includegraphics[width=0.4\textwidth]{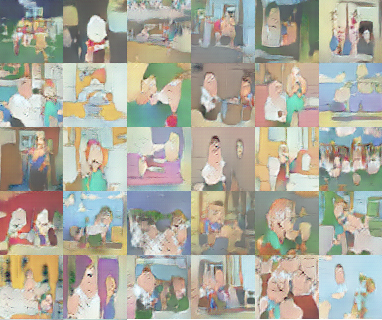} \\ & & & \\[-8pt] \midrule
					& & & \\ 
					\begin{sideways}{~\hspace{1in} \textsf{$128 \times 128$}}\end{sideways} &
					\includegraphics[width=0.4\textwidth]{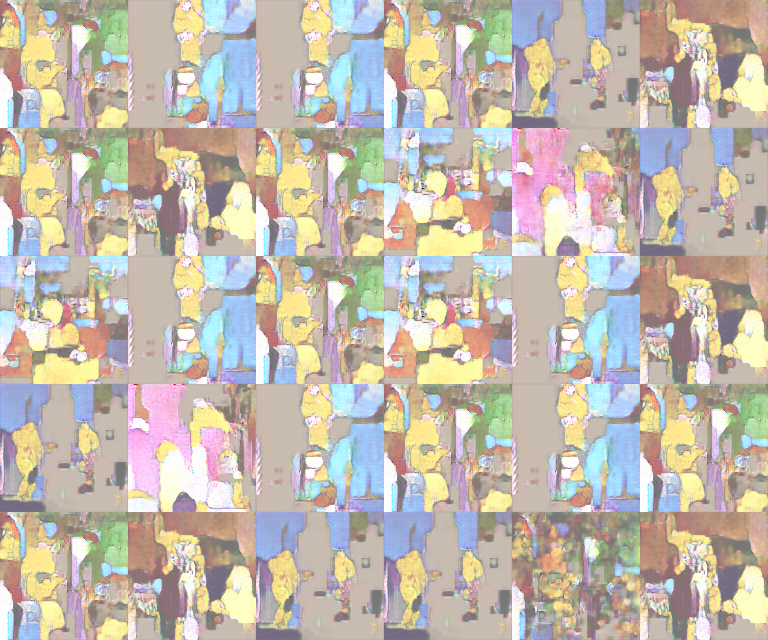} &&&
					\includegraphics[width=0.4\textwidth]{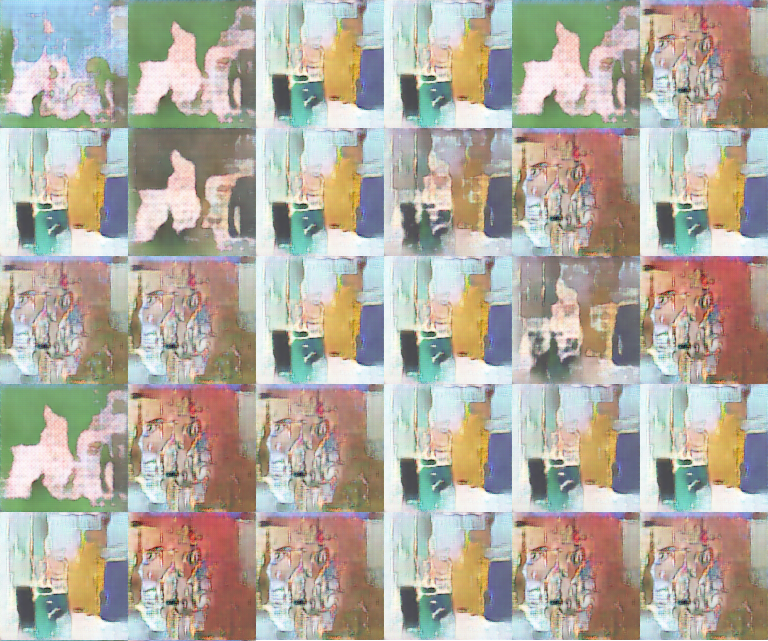} \\
					\bottomrule
				\end{tabular}
			\end{center}
		}}
		\vspace{-10pt}
		\caption{Sample frames generated by two independently-trained cartoon series models, at two different resolutions.}
		\label{fig:single}
	\end{figure*}
	
	We now present results  on generating images
	across our two domains of interest: frames from the cartoon series The Simpsons and Family Guy.
	
	\subsection{Datasets}
	
	To build a large-scale dataset, we took about 25 hours of DVDs
	corresponding to every episode for four years of each program (Simpsons seasons 7-10, Family Guy seasons 1-4).
	We performed screen captures at about
	$\frac{1}{3}$Hz and a resolution of 1280 $\times$ 800, yielding about 400 snapshots per episode or about
	30,000 frames per series in total.
	
	\subsection{Single domain training }

	We began by training a separate, independent network for each of our
	two domains, basing the network architecture and training procedure on
	the publicly-available Torch implementation by Radford \textit{et
		al.}~\cite{radford2015unsupervised}. We generated new $64 \times 64$
	pixel frames by passing independently-sampled random vectors into the
	network's 1024D $z$ input.  Figure~\ref{fig:single}(a) presents the
	results.  Comparing these novel frames to those in the training set
	(Figure~\ref{fig:Intro_fg}), we observe that the network seems to have
	captured the overall style and appearance of the characteristics of
	the original domains --- the distinctive yellow color of the
	characters in the Simpsons, for example, versus the paler skin
	tones in Family Guy.

	\newcommand{\knn}[1]{
		\setlength\fboxsep{0pt}
		\begin{minipage}{1.5cm}
			{{{\includegraphics[width=\textwidth,clip,trim=0cm 2.25cm 11.25cm 0cm]{#1}}}}
		\end{minipage}
		\begin{minipage}{4cm}
			\includegraphics[width=\textwidth,clip,trim=4.5cm 2.25cm 0cm 0cm]{#1} 
			\includegraphics[width=\textwidth,clip,trim=0cm 0cm 4.5cm 2.25cm]{#1}
		\end{minipage}}
		
		\begin{figure*}[t]
			\begin{center}
				{\footnotesize{
						\begin{tabular}{@{}c@{\,}c@{\,\,}|@{\,}c@{\,\,}|@{\,}c@{}} \toprule
							\multirow{3}{*}[-4em]{\begin{sideways}{{\textsf{Simpsons}\,\,\,\,\,\,\,\,\,\,\,\,\,\,\,\,\,}}\end{sideways}} &
							\knn{figures/smp64/knn/Res_clust_8_vis_395.png} &
							\knn{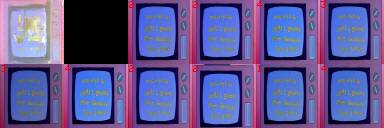} &
							\knn{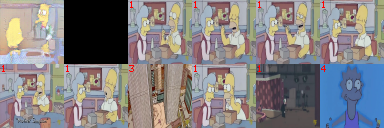} \\[28pt] \cline{2-4}
							& & & \\[-4pt]  
							& 	\knn{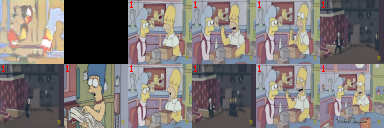} &
							\knn{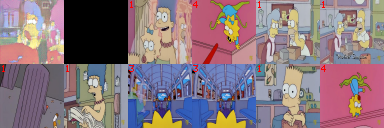} & 
							\knn{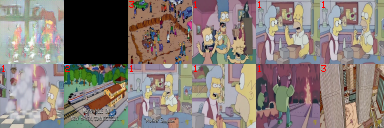} \\[28pt] \cline{2-4}
							& & & \\[-4pt]  
							&	\knn{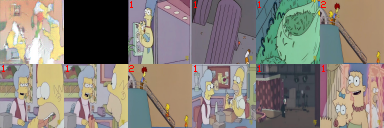} &
							\knn{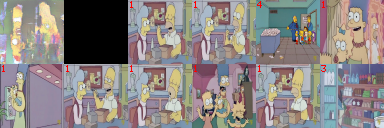} &
							\knn{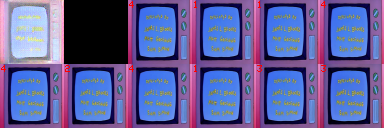} \\[28pt] \midrule
							& & & \\[-4pt]  
							\multirow{4}{*}[-4em]{\begin{sideways}{{\textsf{Family Guy}\,\,\,\,\,\,\,\,\,\,\,\,}}\end{sideways}} &
							\knn{figures/fg64/knn/Res_clust_9_vis_787.png} &
							\knn{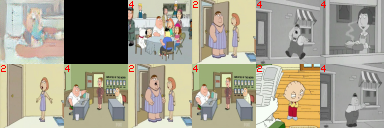} &
							\knn{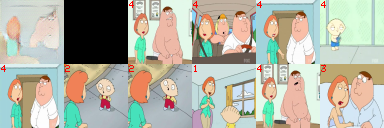} \\[28pt] \cline{2-4}
							& & & \\[-4pt]  
							&	\knn{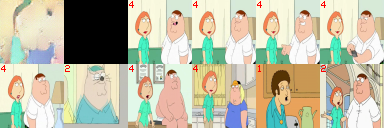} & 
							\knn{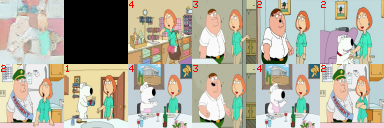} &
							\knn{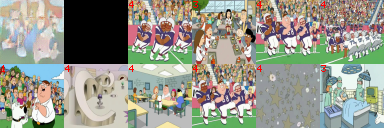}\\[28pt] \cline{2-4}
							& & & \\[-4pt]  	
							&  \knn{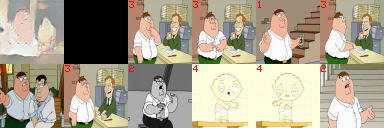} &
							\knn{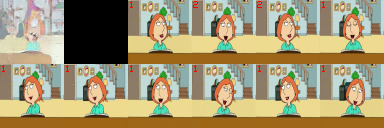} &
							\knn{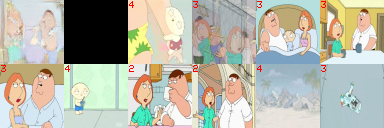} \\[10pt]	
							\bottomrule
						\end{tabular}
					}}
				\end{center}
				\vspace{-10pt}
				\caption{Nearest training neighbors for each of nine sample generated images (left image in each pane) for a GAN model trained on each independent cartoon series.}
				\label{fig:simpsonfamily64_knn}
			\end{figure*}
			
			\xhdr{Nearest neighbors.} To what extent are the generated frames really
			novel, and to what extent do they simply ``copy'' the training images?
			To help visualize this, 
			for each generated frame we found the nearest-neighbors in
			the training set, according to Euclidean
			distance between the activation values of the second to the last layer
			of the discriminator.  Figure~\ref{fig:simpsonfamily64_knn} presents
			the results. Intuitively, the nearest neighbors give us some information
			about the ``inspiration'' that the network used in generating a new frame.
			We observe, for example, that 
			the upper-left image in the
			Family Guy row looks like an image of the husband
			and wife talking in the bedroom,
			but the closest images retrieved from the training dataset are images of them talking in
			the kitchen; this is explained by the fact that the two rooms have very similar
			appearance in the series. 
			In the top-middle example of the Simpsons row, the network appears to have generated a novel
			scene with some people appearing in a TV set, whereas the closest frames in the training
			images are the TV with different varieties of text.
			
			\xhdr{Generating higher-resolution images.}
			To improve the quality of the frames, we tried to generate images at a
			higher resolution of $128 \times 128$, as shown in
			Figure~\ref{fig:single}(b) using the network architecture of
			Figure~\ref{fig:net_all}(c).  Ironically, this network produces frames
			that are subjectively worse: the resolution is higher, but the model
			seems not to have learned the important properties of the source
			domain, and diversity of generated samples is low.  This seems to
			occur because the network has reached a model collapse case where it
			has learned to generate nearly identical images.
			%It may be possible to
			%generate better images with more complex network architectures.

			%An obvious criticism of our generated frames is that they are
			%relatively low resolution. We tried to generate higher resolution
			%images at $128 \times 128$, as shown in the second row of
			%Figure~\ref{fig:single}.  While the images for Simpsons look
			%relatively good, the Family Guy images do not. \djc{Eman, why is
			%this? I am not sure that I said the last statement,(While the images
			%for Simpsons look relatively good, the Family Guy images) both looks
			%not good to me ... I couldn't generate a better higher resolution
			%images based on my trials, I am not sure that it is not doable, I
			%didn't try many things to enhance the resolution, like adding
			%different loss function, or making a generator network like Unet
			%architecture or encode-decoder style with helps improve the training,
			%but I am only become acquainted with these as part of GAN networks
			%when I was already done with the project, even wrote the first
			%version of the paper }
			
			\subsection{Coupled Domain Training }
			
			Instead of training models for the two domains in isolation, we next
			consider various jointly-trained models. We hypothesize that such
			jointly-trained models could potentially overcome the problems with
			generating higher-resolution images (by doubling  the number
			of training examples), and could find semantic correspondences between frames
			across the two series.

			\begin{figure*}[phtb]
				\vspace{1in}
				\begin{center}
					\begin{tabular}{p{12pt}@{}cc|cc} 
						& {$64 \times 64$} &\multicolumn{2}{c}{}& {$128 \times 128$} \\ 
						\toprule
						&
						\includegraphics[width=0.44\textwidth]{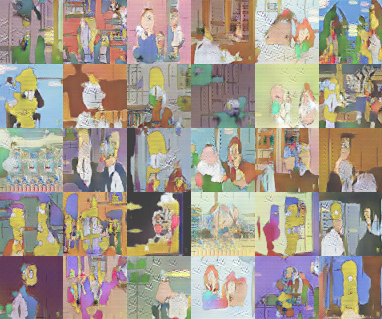} &&&
						\includegraphics[width=0.44\textwidth]{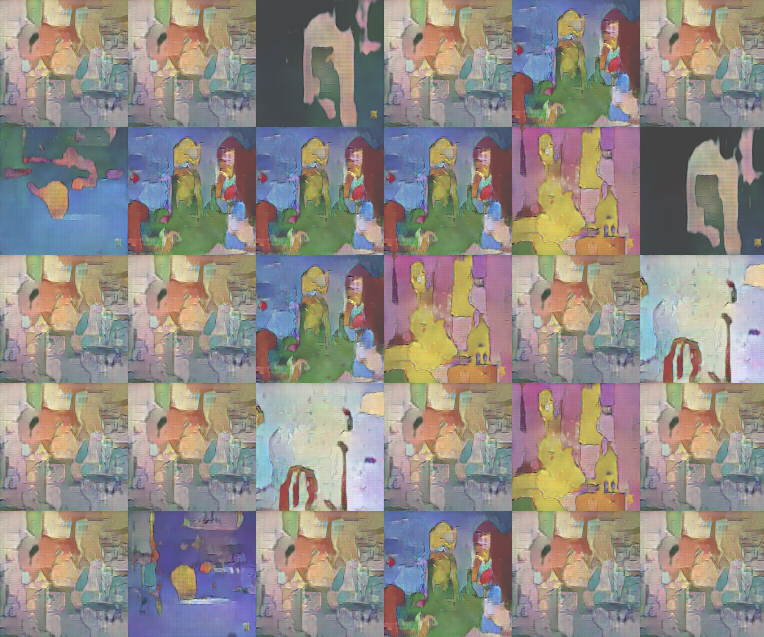} \\ 
						\bottomrule
					\end{tabular} 
					\vspace{1pt} \\
					(a) Frames generated by a single network trained with a mixture of Simpsons and Family Guy frames, at two resolutions.
					
					\vspace{18pt}
					
					\begin{tabular}{c@{\,\,}|@{\,}c@{\,\,}|@{\,}c@{}} \toprule
						\knn{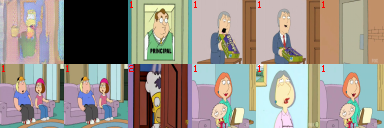} &
						\knn{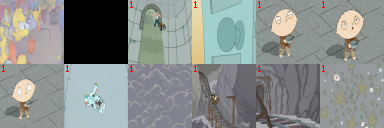} &
						\knn{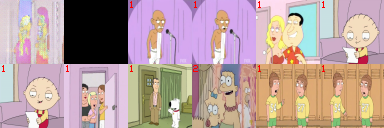} \\[26pt] \hline
						& & \\[-8pt]
						\knn{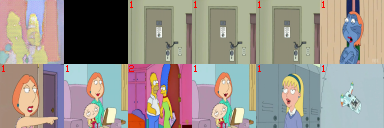} &
						\knn{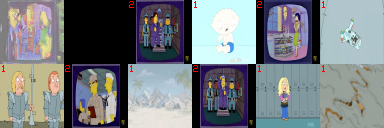} &
						\knn{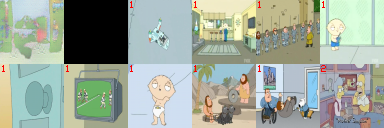}\\[26pt]   \hline
						& & \\[-8pt]
						\knn{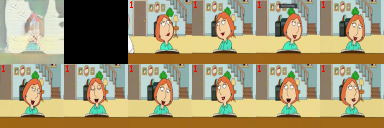} &
						\knn{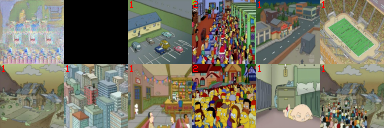} &
						\knn{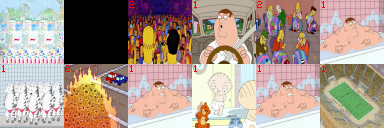}\\     \bottomrule 
					\end{tabular}
					\vspace{1pt} \\
					(b) Nearest training neighbors for each of nine sample frames (left image in each pane) generated by the combined model.
				\end{center}
				\vspace{-10pt}
				\caption{Sample results from a single network trained on an unstructured (unlabeled) mix of the Simpsons and Family Guy data, showing (a) sample generated frames, and (b) nearest training set neighbors for some sample generated frames.}
				\label{fig:combined}
				\vspace{1in}
			\end{figure*}

			\xhdr{Combining the datasets.}  
			We first tried retraining the same
			model as before but with a dataset consisting of both series mixed together,
			and results are in Figure~\ref{fig:combined} for two different resolutions.
			Again, the lower-resolution images seem reasonable in both appearance
			and diversity, and for most frames we can identify characteristics of one or both of 
			the two series. However, the combined dataset seems
			not to have helped the higher-resolution
			frame generator, as we still see evidence of model collapse.
			The second row of the figure again presents $k$-nearest-neighbors in
			the combined training set, showing that sometimes the generated frames 
			are most similar to one dataset, and sometimes seem to synthesize a combination
			of the two.

			%In this section, we examine the results generated from a network trained on
			%both domains jointly using the CO-GAN and domain adaptation
			%models discussed above. 
			
			%The following subsections will examine whether any of the joint
			%training variations has managed to find high level semantic
			%correspondence between the two domains or not, as well as whether that
			%high level semantic correspondence is ambiguous term and the function
			%mapping between the two domain is many-to-many (not bijective) so we
			%may need other modalities like text to create bi-jective mapping.
			
			\begin{figure*}[t]
				\vspace{1in}
				\begin{center}
					\small{
						\begin{tabular}{p{12pt}@{}cc|cc} 
							& \textsf{The Simpsons} &\multicolumn{2}{c}{}& \textsf{Family Guy} \\ 
							\toprule
							\begin{sideways}{~\hspace{0.8in}\textsf{$64 \times 64$}}\end{sideways} &
							\includegraphics[width=0.35\textwidth]{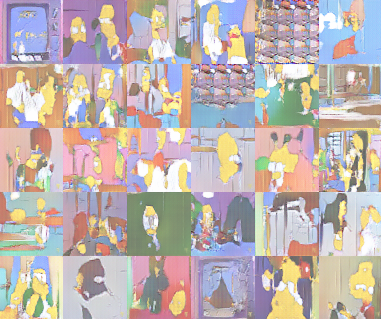} &&&
							\includegraphics[width=0.35\textwidth]{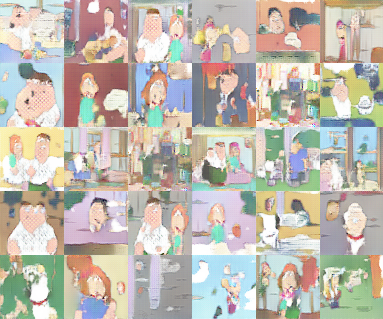} \\ \midrule
							\begin{sideways}{~\hspace{0.8in} \textsf{$128 \times 128$}}\end{sideways} &
							\includegraphics[width=0.35\textwidth]{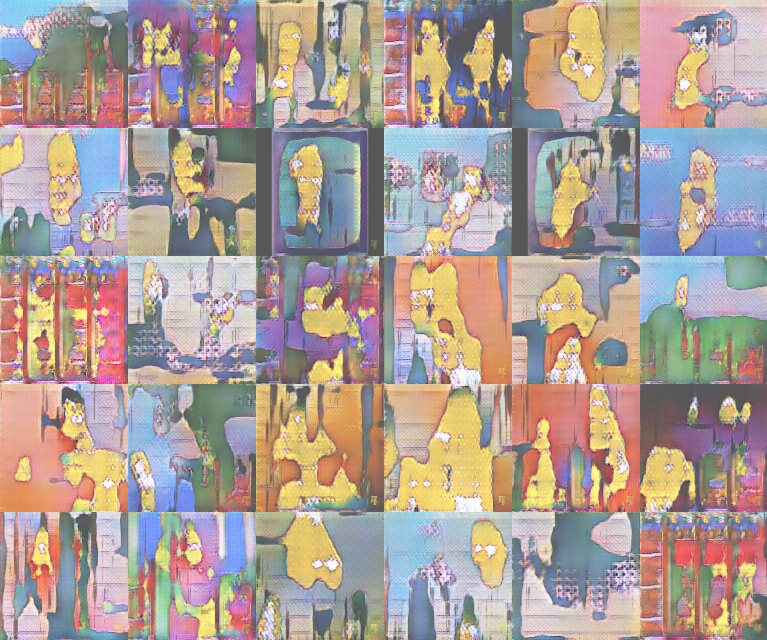} &&&
							\includegraphics[width=0.35\textwidth]{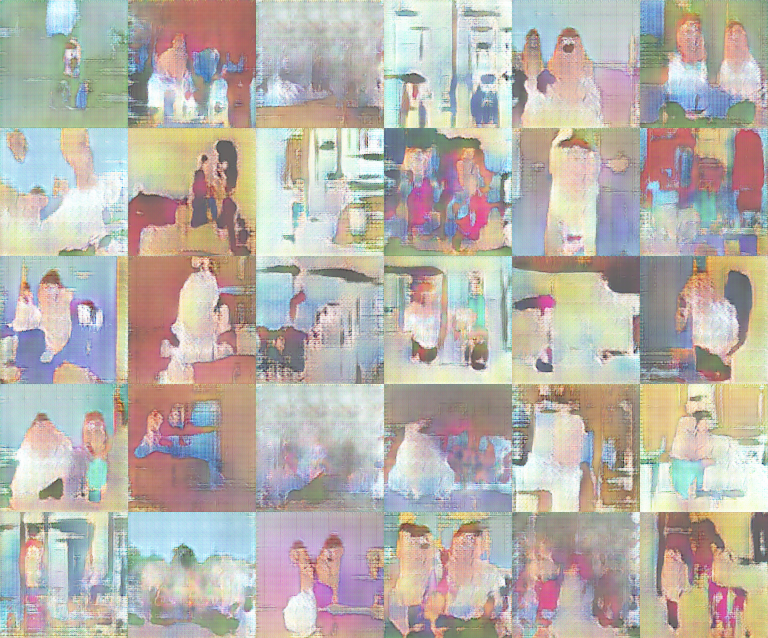} \\ \midrule
							\begin{sideways}{~\hspace{0.8in} \textsf{$256 \times 256$}}\end{sideways} &
							\includegraphics[width=0.35\textwidth]{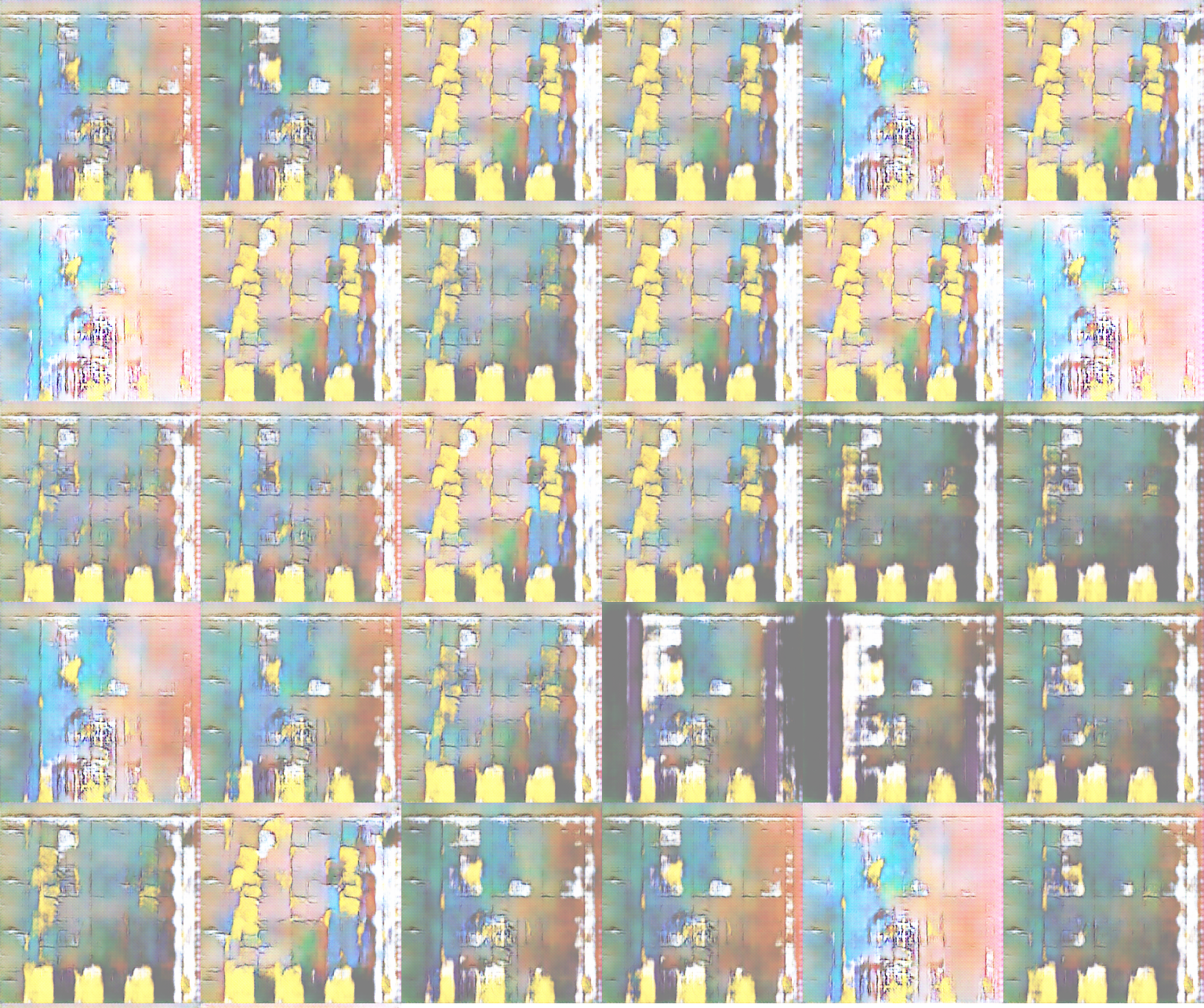} &&&
							\includegraphics[width=0.35\textwidth]{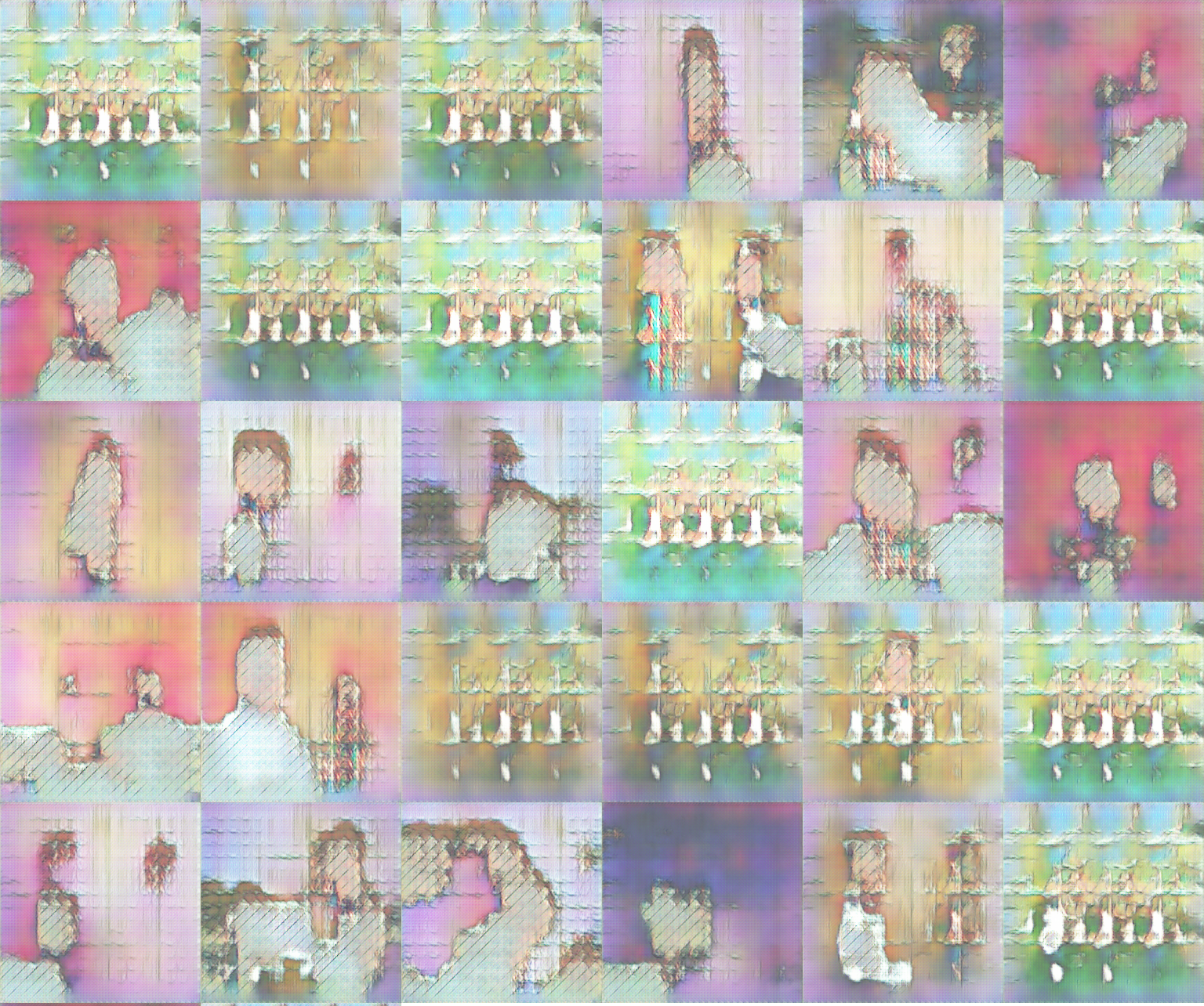} \\ \bottomrule
						\end{tabular}
					}
				\end{center}
				\vspace{-10pt}
				\caption{Sample frames generated by COGAN at different resolutions (rows) for each cartoon dataset (columns).}
				\label{fig:cogan_all_samples}
				\vspace{0.7in}
			\end{figure*}

			\xhdr{COGANs.} We next test COGANs, which explicitly model that there
			are multiple image domains, as shown in 
			Figure~\ref{fig:cogan_all_samples}.
			We observe
			that the coupled training generated images with noticeably better
			quality at $128\times128$, compared to the results without
			coupled training (Figure~\ref{fig:single}) or the single model with combined training datasets (Figure~\ref{fig:combined}).

			\newcommand{\knnpair}[1]{
				\setlength\fboxsep{0pt}
				\begin{minipage}{2cm}
					{{{\includegraphics[width=\textwidth,clip,trim=0cm 2.25cm 11.25cm 4.5cm]{#1}}}} \\
					{{{\includegraphics[width=\textwidth,clip,trim=0cm 6.75cm 11.25cm 0cm]{#1}}}} 
				\end{minipage}
				\begin{minipage}{4cm}
					\includegraphics[width=\textwidth,clip,trim=4.5cm 2.25cm 0cm 4.5cm]{#1}  
					\includegraphics[width=\textwidth,clip,trim=0cm 0cm 4.5cm 6.75cm]{#1} \\
					\includegraphics[width=\textwidth,clip,trim=4.5cm 6.75cm 0cm 0cm]{#1}  
					\includegraphics[width=\textwidth,clip,trim=0cm 4.5cm 4.5cm 2.25cm]{#1} 
				\end{minipage}}
				
				\newcommand{\coganheading}{\multirow{1}{*}[4em]{\begin{sideways} \textsf{\scriptsize{\,\,\,\,\,\,\,\,\,\,\,Family Guy \,\,\,\,\,\,\,\,\,\,\,\,\,\,\,\,\,\,\, Simpsons}}  \end{sideways}} &}
				
				\begin{figure*}[t]
					\vspace{1in}
					\begin{center}
						\resizebox{1.0\textwidth}{!}{
							\begin{tabular}{@{}p{12pt}@{}c|c|c@{}} \toprule
								\coganheading
								\knnpair{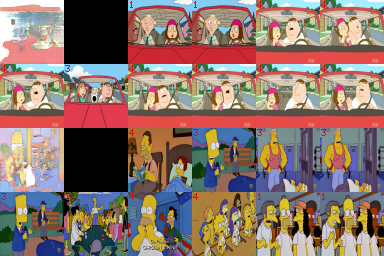}&
								\knnpair{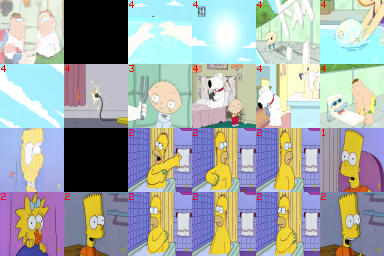} &
								\knnpair{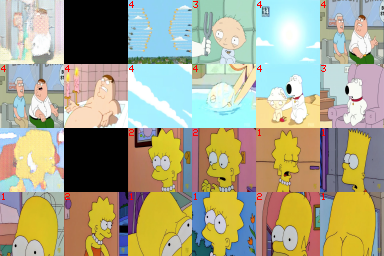} \\ 
								&
								& & \\[-8pt] \hline
								&	& & \\[-8pt] 
								\coganheading
								\knnpair{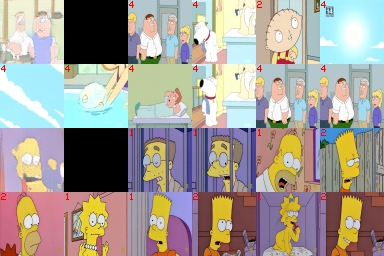} &
								\knnpair{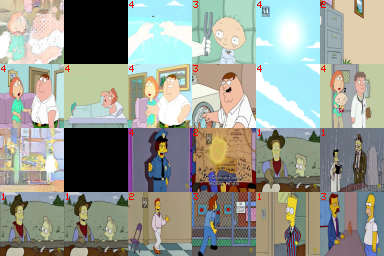}  &
								\knnpair{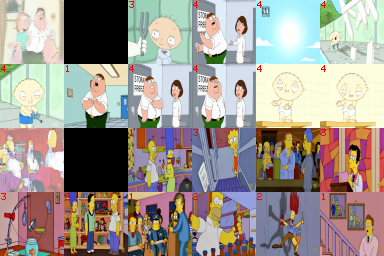}\\ 
								&
								& & \\[-8pt] \hline
								&	& & \\[-8pt] 
								\coganheading
								\knnpair{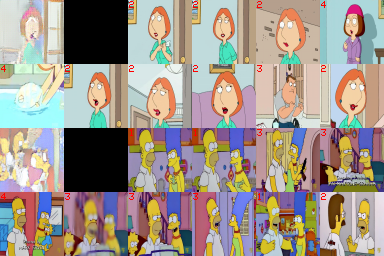} &
								\knnpair{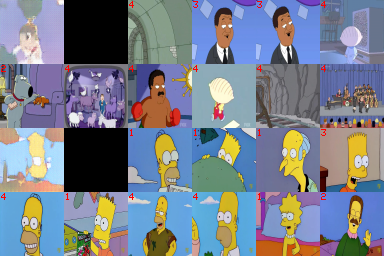} &
								\knnpair{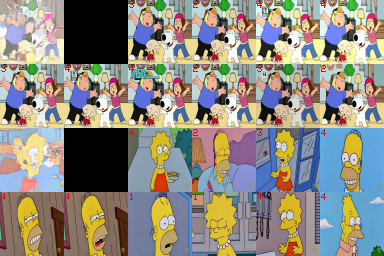}\\  
								&
								& & \\[-8pt] \hline
								&	& & \\[-8pt] 
								\coganheading
								\knnpair{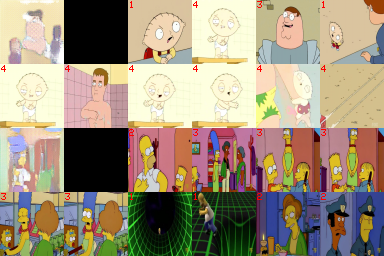} &
								\knnpair{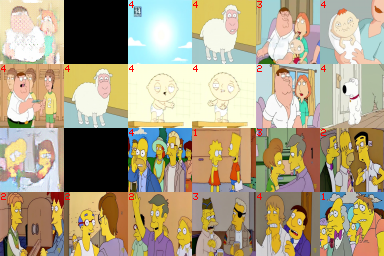} &
								\knnpair{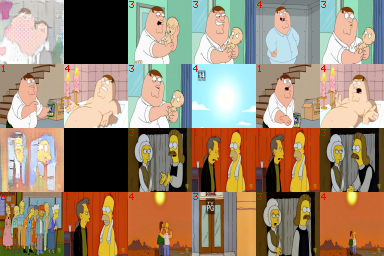}\\  \bottomrule
							\end{tabular}
						}
					\end{center}
					\vspace{-8pt}
					\caption{Sample results of paired image generation with COGAN. In each pane, the top left and bottom right images were generated
						by the same random input vector passed to the Simpsons and Family Guy models, respectively. The remaining images show nearest neighbors
						in the corresponding dataset.}
					\label{fig:cogan_knn_64}
					\vspace{1in}
				\end{figure*}
				
				%Each image in the figure contains four rows,
				%the first two rows represents Family Guy results while the second two
				%rows represent Simpson results. 

				%% Then similarly we display the K-NN results with $K=10$. But this time
				%% we can have two ways of feature extraction. The first one is from the
				%% discriminator as the previous techniques. So for each generated image
				%% we retrieve its corresponding K-nearest neighbor from the same domain
				%% and the opposite domain. In the second technique, we get the features
				%% from the classifier modules instead of from the discriminator and use
				%% the feature vectors to search for the nearest K-NN with $k=10$ from
				%% the same domain and from the opposite domain and see whether there is
				%% a correspondence between the two methods or not.
				
				%% The results of this technique are shown in Figures
				%% \ref{fig:DA_knn_64},\ref{fig:DA_knn_64_v1}, \ref{fig:DA_knn_64_v3} and
				%% \ref{fig:DA_knn_64_v4}. The figure structure is as follows: the first
				%% column shows the results generated by the first method and the second
				%% column shows results generated by the second method. The other two
				%% columns are just more results, where each subfigure consists of sets
				%% of images showing the KNN for each of its same domain or the
				%% corresponding domain, so we may be able to detect some high level
				%% semantic correspondences. The results of the main algorithm shown in
				%% Figure \ref{fig:DA_knn_64}.

				\newcounter{lblno}
				\newcounter{frameno}
				%\DeclareRobustCommand{\lbllet}{{{\stepcounter{lblno}\thelblno}}}

				%\newcommand{\lbl}[1]{\multicolumn{1}{r}{\textbf{\texttt{{\textcolor{black}{\contour{white}{#1}}}}}}}
				%\newcommand{\lbl}{\stepcounter{lblno}{\textcolor{black}{\fontsize{5.5}{4}\selectfont{\texttt{\contour{white}{\,\,\,\alph{lblno}}}}}}}
				\newcommand{\lblframe}{\stepcounter{frameno}{\textcolor{black}{\fontsize{9}{9}\selectfont{\texttt{\contour{white}{\theframeno}}}}}}
				\newcommand{\lbl}[1]{\stepcounter{lblno}{\textcolor{black}{\fontsize{5.5}{4}\selectfont{\texttt{\contour{white}{#1}}}}}}
				\newcommand{\labels}[5]{
					\lbl{#1} \hspace{13.5pt}
					\lbl{#2} \hspace{13.5pt}
					\lbl{#3} \hspace{13.5pt}
					\lbl{#4} \hspace{13.5pt}
					\lbl{#5} \hspace{13.5pt}
				}
				\newcommand{\knndaImgsHelper}[6]{
					\setlength\fboxsep{0pt}
					#1 &
					\begin{minipage}{1.55cm}
						{{{\includegraphics[width=0.95\textwidth,clip,trim=0cm 6.75cm 11.3cm 0cm]{#2}}}}
					\end{minipage} &
					\begin{minipage}{3.6cm}
						\includegraphics[width=1\textwidth]{#3}  
					\end{minipage} &
					\begin{minipage}{3.6cm}
						\includegraphics[width=1\textwidth]{#4}  
					\end{minipage} &
					\begin{minipage}{3.6cm}
						\includegraphics[width=1\textwidth]{#5}  
					\end{minipage} &
					\begin{minipage}{3.6cm}
						\includegraphics[width=1\textwidth]{#6}  
					\end{minipage}\\[-30pt]
					& & \setcounter{lblno}{0}
					%\labels{a}{b}{c}{d}{e} & \labels{k}{l}{m}{n}{o} & \labels{\textalpha}{\textbeta}{\textgamma}{\textdelta}{\textepsilon} & \labels{\textlambda}{\textmu}{\textnu}{\textxi}{\textomikron} \\[11pt]
					\labels{a}{b}{c}{d}{e} & \labels{k}{l}{m}{n}{o} & \labels{A}{B}{C}{D}{E} & \labels{K}{L}{M}{N}{O} \\[11pt]
					& & 
					%\labels{f}{g}{h}{i}{j} &\labels{p}{q}{r}{s}{t} & \labels{\textzeta}{\texteta}{\texttheta}{\textiota}{\textkappa} & \labels{\textpi}{\textrho}{\textsigma}{\texttau}{\textupsilon} \\[-40pt]
					\labels{f}{g}{h}{i}{j} &\labels{p}{q}{r}{s}{t} & \labels{F}{G}{H}{I}{J} & \labels{P}{Q}{R}{S}{T} \\[-40pt]
					& \hspace{2pt}\lblframe & & & \\[32pt]
					%& F1 & F2 & {\multicolumn{1}{r}{{{\textsf{F3}}}}} & F4 & F5  \\[3pt]
				}

				\newcommand{\knndaImgsInit}[6]{
					\setcounter{lblno}{0}
					\knndaImgsHelper{{\multirow{16}{*}[-1em]{\begin{sideways}{\textsf{#1}}\end{sideways}}}}{#2}{#3}{#4}{#5}{#6}
					\cmidrule{2-6}
				}
				
				\newcommand{\knndaImgs}[5]{
					\knndaImgsHelper{}{#1}{#2}{#3}{#4}{#5}
					\cmidrule{2-6}
				}
				
				\newcommand{\knndaImgsLast}[5]{
					\knndaImgsHelper{}{#1}{#2}{#3}{#4}{#5}
				}

				\newcommand{\knnda}[2]{
					\setlength\fboxsep{0pt}
					\begin{minipage}{1.75cm}
						{{{\includegraphics[width=\textwidth,clip,trim=0cm 6.75cm 11.25cm 0cm]{#1}}}}
					\end{minipage} &
					\begin{minipage}{3.5cm}
						\noindent
						\includegraphics[width=0.77\textwidth,clip,trim=4.5cm 6.75cm 0cm 0cm]{#1} 
						\noindent
						\includegraphics[width=0.18\textwidth,clip,trim=11.25cm 4.5cm 0cm 2.25cm]{#1}
						\noindent
						\includegraphics[width=0.95\textwidth,clip,trim=0cm 4.5cm 2.25cm 2.25cm]{#1} 
					\end{minipage} &
					\begin{minipage}{3.5cm}
						\includegraphics[width=0.77\textwidth,clip,trim=4.5cm 2.25cm 0cm 4.5cm]{#1}  
						\includegraphics[width=0.195\textwidth,clip,trim=11.25cm 0cm 0cm 6.75cm]{#1}  
						\includegraphics[width=0.95\textwidth,clip,trim=0cm 0cm 2.25cm 6.75cm]{#1} 
					\end{minipage} &
					\begin{minipage}{3.5cm}
						\includegraphics[width=0.77\textwidth,clip,trim=4.5cm 6.75cm 0cm 0cm]{#2}  
						\includegraphics[width=0.195\textwidth,clip,trim=11.25cm 4.5cm 0cm 2.25cm]{#2}
						\includegraphics[width=0.95\textwidth,clip,trim=0cm 4.5cm 2.25cm 2.25cm]{#2} 
					\end{minipage} &
					\begin{minipage}{3.5cm}
						\includegraphics[width=0.77\textwidth,clip,trim=4.5cm 2.25cm 0cm 4.5cm]{#2}  
						\includegraphics[width=0.195\textwidth,clip,trim=11.25cm 0cm 0cm 6.75cm]{#2}  
						\includegraphics[width=0.95\textwidth,clip,trim=0cm 0cm 2.25cm 6.75cm]{#2} 
					\end{minipage}\\
				}
				
				\newcommand{\knndarev}[2]{
					\setlength\fboxsep{0pt}
					\begin{minipage}{1.75cm}
						{{{\includegraphics[width=\textwidth,clip,trim=0cm 6.75cm 11.25cm 0cm]{#1}}}}
					\end{minipage} &
					\begin{minipage}{2.6cm}
						\includegraphics[width=\textwidth,clip,trim=4.5cm 2.25cm 2.25cm 4.5cm]{#1}  
						\includegraphics[width=0.329\textwidth,clip,trim=11.25cm 2.25cm 0cm 4.5cm]{#1}  
						\includegraphics[width=0.65\textwidth,clip,trim=0cm 0cm 9cm 6.75cm]{#1} 
					\end{minipage} &
					\begin{minipage}{2.6cm}
						\includegraphics[width=\textwidth,clip,trim=4.5cm 6.75cm 2.25cm 0cm]{#1}  
						\includegraphics[width=0.329\textwidth,clip,trim=11.25cm 6.75cm 0cm 0cm]{#1}  
						\includegraphics[width=0.65\textwidth,clip,trim=0cm 4.5cm 9cm 2.25cm]{#1} 
					\end{minipage} &
					\begin{minipage}{2.6cm}
						\includegraphics[width=\textwidth,clip,trim=4.5cm 2.25cm 2.25cm 4.5cm]{#2}  
						\includegraphics[width=0.329\textwidth,clip,trim=11.25cm 2.25cm 0cm 4.5cm]{#2}  
						\includegraphics[width=0.65\textwidth,clip,trim=0cm 0cm 9cm 6.75cm]{#2} 
					\end{minipage} &
					\begin{minipage}{2.6cm}
						\includegraphics[width=\textwidth,clip,trim=4.5cm 6.75cm 2.25cm 0cm]{#2}  
						\includegraphics[width=0.329\textwidth,clip,trim=11.25cm 6.75cm 0cm 0cm]{#2}  
						\includegraphics[width=0.65\textwidth,clip,trim=0cm 4.5cm 9cm 2.25cm]{#2} 
					\end{minipage}
				}
				
				\newcommand{\knndatwobyfour}[2]{
					\setlength\fboxsep{0pt}
					\begin{minipage}{1.75cm}
						{{{\includegraphics[width=\textwidth,clip,trim=0cm 6.75cm 11.25cm 0cm]{#1}}}} 
					\end{minipage} 
					\begin{minipage}{3.5cm}
						\includegraphics[width=\textwidth,clip,trim=4.5cm 6.75cm 0cm 0cm]{#1}  
						\includegraphics[width=\textwidth,clip,trim=0cm 4.5cm 4.5cm 2.25cm]{#1} 
					\end{minipage}
					\begin{minipage}{3.5cm}
						\includegraphics[width=\textwidth,clip,trim=4.5cm 2.25cm 0cm 4.5cm]{#1}  
						\includegraphics[width=\textwidth,clip,trim=0cm 0cm 4.5cm 6.75cm]{#1} 
					\end{minipage}
					\begin{minipage}{3.5cm}
						\includegraphics[width=\textwidth,clip,trim=4.5cm 6.75cm 0cm 0cm]{#2}  
						\includegraphics[width=\textwidth,clip,trim=0cm 4.5cm 4.5cm 2.25cm]{#2} 
					\end{minipage}
					\begin{minipage}{3.5cm}
						\includegraphics[width=\textwidth,clip,trim=4.5cm 2.25cm 0cm 4.5cm]{#2}  
						\includegraphics[width=\textwidth,clip,trim=0cm 0cm 4.5cm 6.75cm]{#2} 
					\end{minipage}
				}
				
				\newcommand{\darow}[7]{
					\multirow{2}{*}[1em]{\begin{sideways}#1\end{sideways}} & 
					\multirow{2}{*}[2.5em]{	\resizebox{1.765cm}{!}{\fullfourbytwo{#2}}} &
					\multirow{2}{*}[2.5em]{	\resizebox{1.765cm}{!}{\fullfourbytwo{#3}}} &	\knnda{{#4}}{{#5}} \\
					& & & 	\knnda{{#6}}{{#7}} 
				}
				\newcommand{\darowrev}[7]{
					\multirow{2}{*}[1em]{\begin{sideways}#1\end{sideways}} & 
					\multirow{2}{*}[2.5em]{	\resizebox{1.765cm}{!}{\fullfourbytwo{#2}}} &
					\multirow{2}{*}[2.5em]{	\resizebox{1.765cm}{!}{\fullfourbytwo{#3}}} &	\knndarev{{#4}}{{#5}} \\
					& & & 	\knndarev{{#6}}{{#7}} 
				}
				
				%%%%%%%%%%%%%%%%%%%%%%%%%%
				
				\newcommand{\knnheading}{
					&  \multicolumn{1}{c}{}               & \multicolumn{2}{c}{\textsf{\textbf{Discriminator similarity}}} & \multicolumn{2}{c}{\textsf{\textbf{Classifier similarity}}} \\
					&  \multicolumn{1}{c}{\textsf{Frame}} & \multicolumn{1}{c}{\textsf{Family Guy}} & \multicolumn{1}{c}{\textsf{Simpsons}} & \multicolumn{1}{c}{\textsf{Family Guy}} &
					\multicolumn{1}{c}{\textsf{Simpsons}} \\ \toprule
				}
				
				\newcommand{\knncaption}[1]{
					\vspace{-20pt}
					\caption{Sample results of the \textbf{#1} model. For each generated frame (left image of each row), we show the 10 nearest neighbors in each of the two training datasets, using two different definitions of similarity.}
				}
				
				\setcounter{frameno}{0}
				\begin{figure*}[t]
					\begin{center}
						{\footnotesize{
								\begin{tabular}{@{}p{12pt}@{}p{1.55cm}@{\,\,}|@{\,\,}p{3.6cm}@{\,\,}|@{\,\,}p{3.6cm}|p{3.6cm}@{\,\,}|@{\,\,}p{3.6cm}@{}}
									\knnheading
									\knndaImgsInit{Family Guy}{{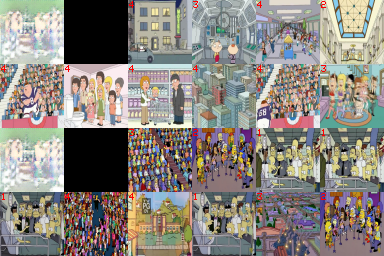}}{{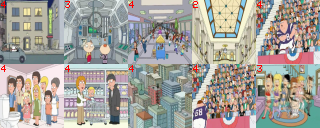}}{{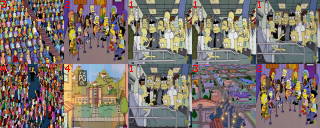}}{{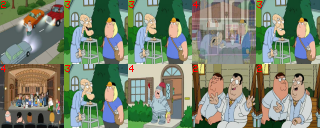}}{{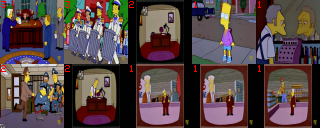}} 
									
									\knndaImgs{{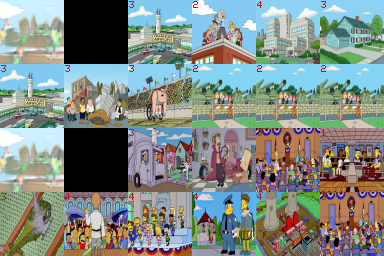}}{{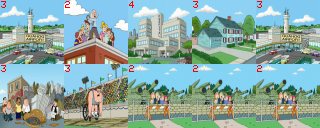}}{{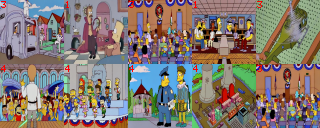}}{{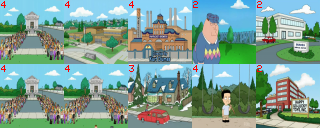}}{{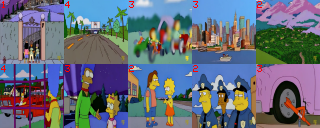}} 
									
									\knndaImgs{{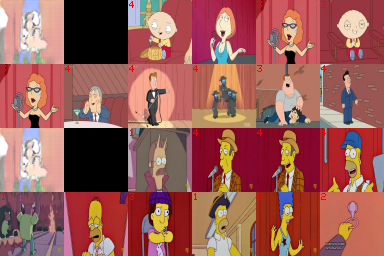}}{{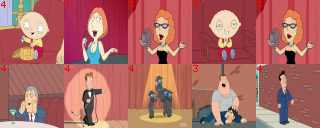}}{{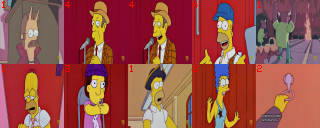}}{{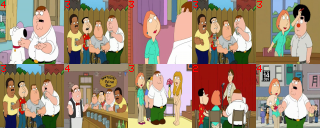}}{{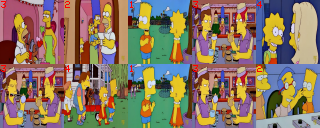}}
									
									\knndaImgsLast{{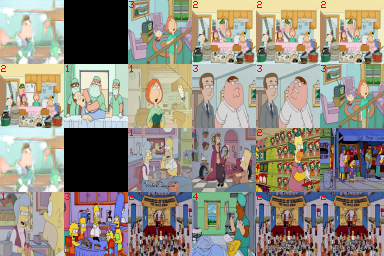}}{{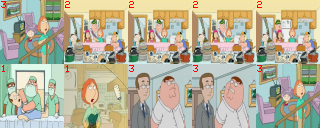}}{{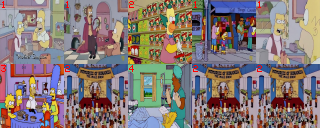}}{{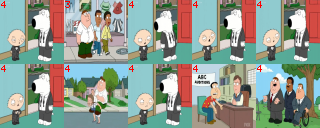}}{{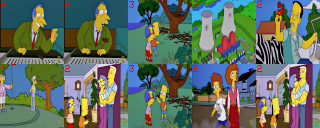}}
									
									\midrule
									
									\knndaImgsInit{Simpsons}{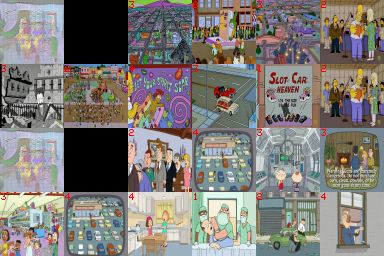}{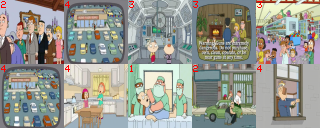}{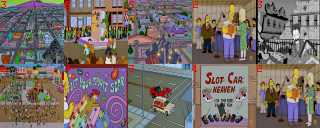}{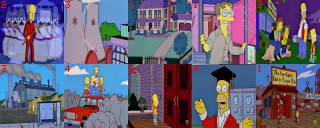}{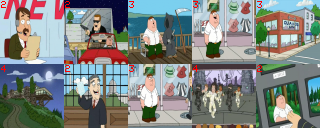}
									
									\knndaImgs{{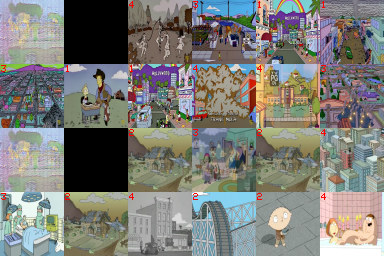}}{{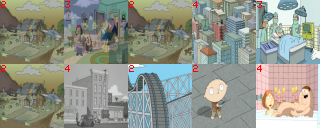}}{{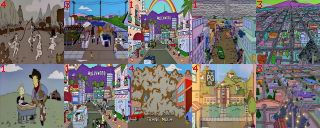}}{{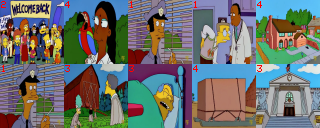}}{{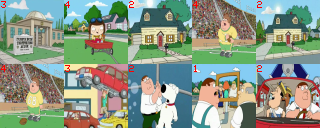}}

									\knndaImgs{{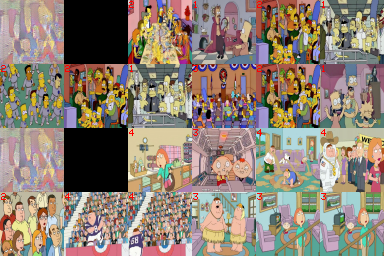}}{{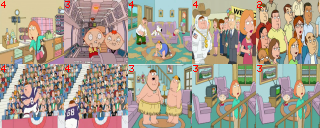}}{{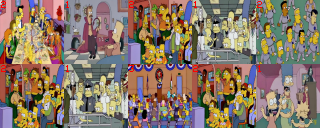}}{{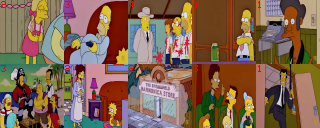}}{{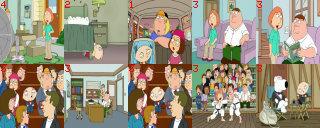}}   
									
									\knndaImgsLast{{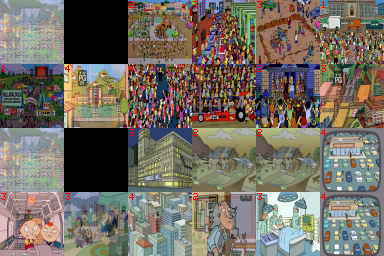}}{{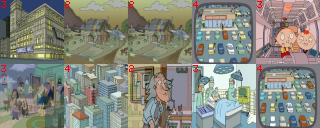}}{{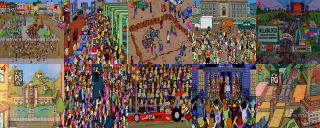}}{{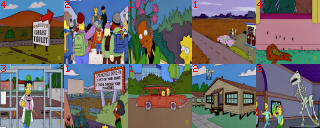}}{{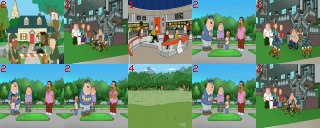}} \midrule
								\end{tabular}
							}}
						\end{center}
						\knncaption{Full Domain Adaptation}
						\label{fig:DA_knn_64}
					\end{figure*}

					How well has the COGAN found and modeled semantic connections between
					the two domains?  To test this, we fed the same random input vectors
					$z$ into both of the two models, as shown in each pane of
					Figure~\ref{fig:cogan_knn_64}. Intuitively, the images generated from
					the same $z$ should be topically similar if COGAN has identified
					meaningful semantic connections; we observe, however that this does
					not appear to be the case, since images across domains in the figure
					are quite different.  We also show the nearest neighbors of each
					generated frame in the corresponding training set.
					%Each image in the figure contains four rows,
					
					\xhdr{Domain adaptation.}  Sample results for domain adaption are
					shown in Figure~\ref{fig:DA_knn_64}.  We noticed during training that
					only some iterations of the model were able to generate good images
					for both domains; here we selected iterations that worked well. 
					The figure also shows the nearest neighbors 
					in each domain's training set for each sample generated
					frame. We use two different features for finding nearest
					neighbors: (1) the second to the last layer of the discriminator, as
					before, and (2) second to the last layer of the classifier. The
					results suggest that the model managed to find correspondences in the
					main colors of the image in both domains. 
					%For the sake on notation consider each nearest neighbor results image as shown in Figure \ref{fig:DA_knn_64} first row and second column. 
					%This image consists of list of $10$ images represented in two rows and five columns. We will use $y_i$ to index rows $i=1,2$ and $x_j$ to represent column where $j=1, \hdots 5$.
					%Results of the whole
					%training system and its different variations are shown in Figures
					%\ref{fig:DA_knn_64}, \ref{fig:DA_knn_64_v1}, \ref{fig:DA_knn_64_v2},
					%\ref{fig:DA_knn_64_v3}, and \ref{fig:DA_knn_64_v4}.
					
					%
					%The results are interesting shown in Figure \ref{fig:DA_knn_64} as in the top example where the algorithm can align images of street cars at rush hours in both domains. Or House in the next two examples. Also it manage to find alignment that represent crowd of people in many examples. Also the seventh example the network find an alignment corresponding to a dominant color
					
					\newcounter{cntresult}
					\newcommand{\freferd}[2]{\def\one{#1}\def\two{#2}\FPeval{\result}{round((\one-1)*5+\two,0)}\setcounter{cntresult}{\result}\alph{cntresult}}
					\newcommand{\freferc}[2]{\def\one{#1}\def\two{#2}\FPeval{\result}{round((\one-1)*5+\two,0)}\setcounter{cntresult}{\result}\Alph{cntresult}}
					\newcommand{\sreferd}[2]{\def\one{#1}\def\two{#2}\FPeval{\result}{round((\one-1)*5+10+\two,0)}\setcounter{cntresult}{\result}\alph{cntresult}}
					\newcommand{\sreferc}[2]{\def\one{#1}\def\two{#2}\FPeval{\result}{round((\one-1)*5+10+\two,0)}\setcounter{cntresult}{\result}\Alph{cntresult}}
					
					Examining the results shown in Figure \ref{fig:DA_knn_64}, we note
					that the model seemed to find some meaningful high-level semantic
					alignments.  For discriminator similarity, for example, we see
					alignments in terms of people in theaters or stadiums (Family Guy
					images 1\freferd{1}{5} and 1\freferd{2}{4} with Simpsons images
					1\sreferd{1}{1} and 1\sreferd{2}{1}), houses (images 2\freferd{1}{1} through 2\freferd{1}{5} with 2\sreferd{1}{1} and 2\sreferd{2}{3}),
					a person talking against a red background (row 3), cars in a parking lot (images 5\freferd{1}{2} and 5\freferd{2}{1} with 5\sreferd{1}{1} and 5\sreferd{1}{3}),
					and groups of people indoors (7\freferd{1}{4} and 7\freferd{1}{5} with 7\sreferd{1}{1} and 7\sreferd{1}{3} through 7\sreferd{2}{5}).
					With the similarity measure in the classifier's feature space, a
					general theme seems to be people conversing in different indoor
					settings, including images 1\freferc{2}{4} and 1\freferc{2}{5} with
					1\sreferc{1}{1}, 1\sreferc{1}{2}, and 1\sreferc{2}{1}.  Images
					2\freferc{1}{4} and 2\freferc{2}{4} seem to relate to image
					2\sreferc{1}{3} in that both have people talking against a green
					background, and in general the green background seems to be a theme of
					the whole row.  The third row appears to roughly correspond with
					conversations between pairs of people, as in images 3\freferc{1}{1},
					3\freferc{1}{3}, and 3\freferc{1}{5} with images
					3\sreferc{1}{3} through 3\sreferc{2}{1}, whereas the fifth row
					features single characters in the scene (e.g.\ images 5\freferc{1}{1},
					5\freferc{1}{4}, and 5\freferc{2}{4} with 5\sreferc{1}{1} through
					5\sreferc{1}{4}).  Other themes include square-framed scenes (images
					1\freferc{2}{1} with 1\sreferc{1}{3} and 1\sreferc{2}{2} through
					1\sreferc{2}{5}) and scenes with prominent buildings (image 6\freferc{1}{5} and
					6\freferc{2}{2} with 6\sreferc{1}{1} through 6\sreferc{1}{3}).

					\begin{figure*}[th]
						{\textsf{\footnotesize{
									\begin{center}
										\begin{tabular}{@{}p{12pt}cc|cc}
											& \multicolumn{1}{c}{The Simpsons} &\multicolumn{2}{@{}c@{}}{} & \multicolumn{1}{@{}c@{}}{Family Guy} \\ \toprule
											\begin{sideways}{\,\,\,\,\,\,\,\,\,Full Domain Adaptation}\end{sideways} &
											\includegraphics[width=0.26\textwidth]{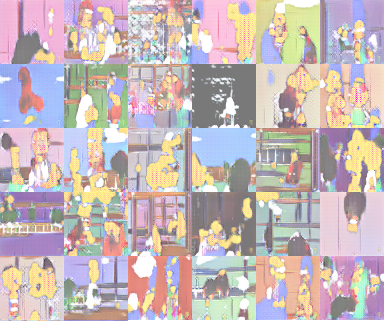} &&&
											\includegraphics[width=0.26\textwidth]{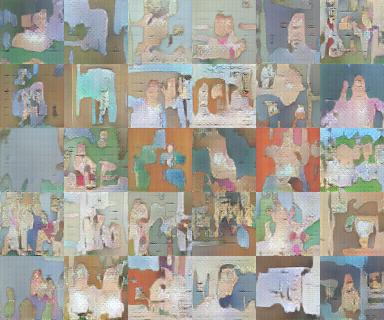} \\ \midrule
											\begin{sideways}{\,\,\,\,\,\,\,\,\,\,\,NoClassifierTraining}\end{sideways} &
											\includegraphics[width=0.26\textwidth]{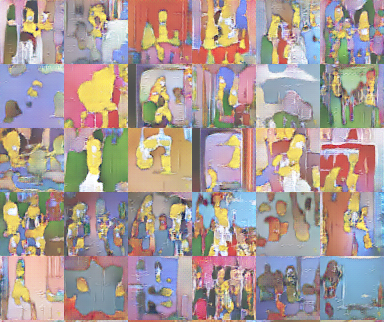} &&&
											\includegraphics[width=0.26\textwidth]{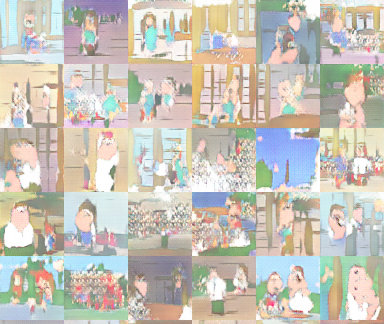} \\ \midrule
											\begin{sideways}{\,\,\,\,\,\,NoFakeClassifierTraining}\end{sideways} &
											\includegraphics[width=0.26\textwidth]{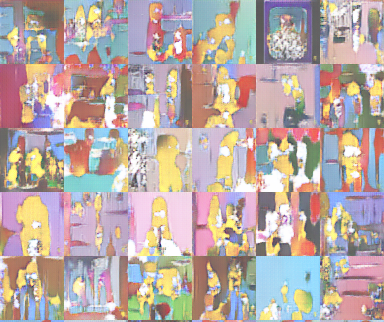} &&&
											\includegraphics[width=0.26\textwidth]{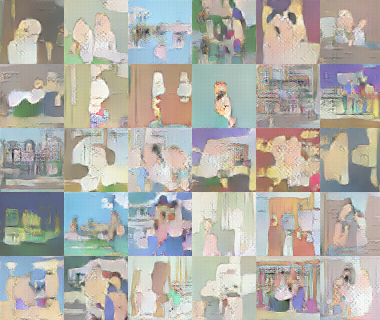} \\ \midrule
											\begin{sideways}{\,\,\,\,\,\,NoRealClassifierTraining}\end{sideways} &
											\includegraphics[width=0.26\textwidth]{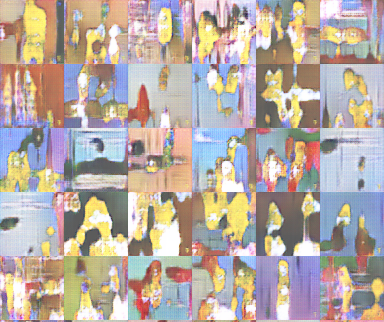} &&&
											\includegraphics[width=0.26\textwidth]{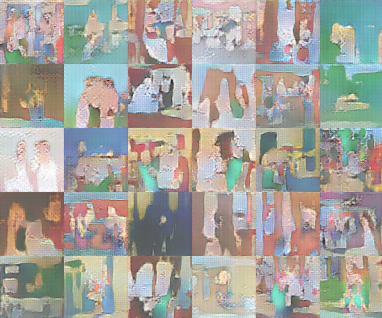} \\ \midrule
											\begin{sideways}{\,\,\,LazyFakeClassifierTraining}\end{sideways} &
											\includegraphics[width=0.26\textwidth]{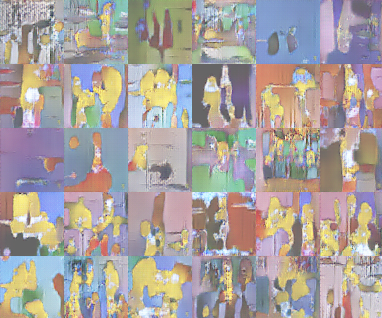} &&&
											\includegraphics[width=0.26\textwidth]{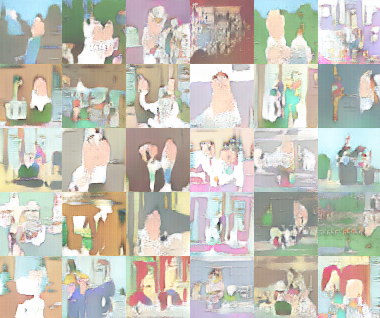} \\ \midrule
										\end{tabular}
									\end{center}
								}}}
								\vspace{-10pt}
								\caption{Sample frames generated under different variants of the domain adaptation models.}
								\label{fig:DA_all_Samples}
							\end{figure*}
							
							%%%%%%%%%%%%%%%%%%%%%%%%%%%%%%%%%%%%%%%%%%%%%%%
							\setcounter{frameno}{0}
							\begin{figure*}[th]
								\begin{center}
									{\footnotesize{
											\begin{tabular}{@{}p{12pt}@{}p{1.55cm}@{\,\,}|@{\,\,}p{3.6cm}@{\,\,}|@{\,\,}p{3.6cm}@{\,\,}|@{\,\,}p{3.6cm}@{\,\,}|@{\,\,}p{3.6cm}@{}}
												\knnheading
												
												\knndaImgsInit{Family Guy}{{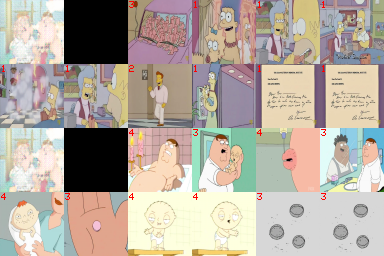}}{{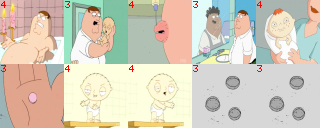}}{{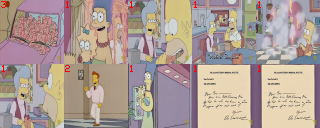}}{{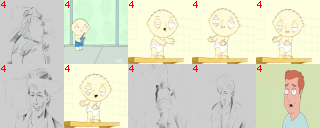}}{{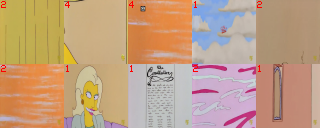}} 
												
												\knndaImgs{{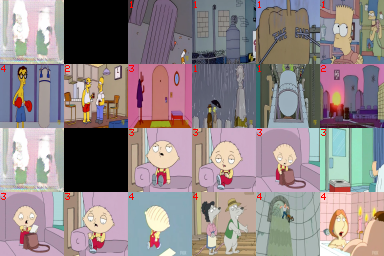}}{{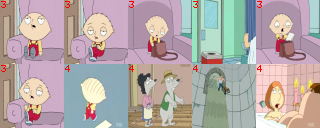}}{{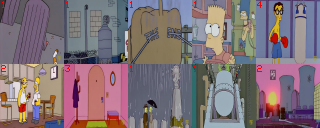}}{{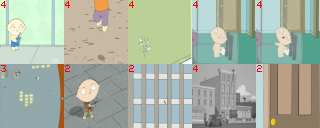}}{{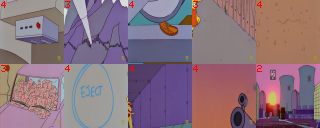}}
												
												\knndaImgs{{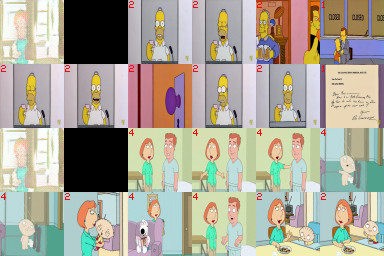}}{{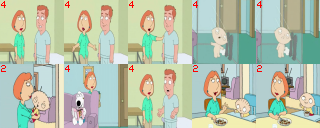}}{{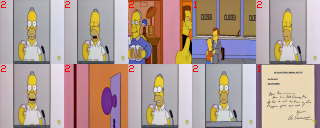}}{{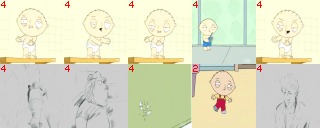}}{{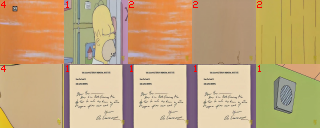}}
												
												\knndaImgsLast{{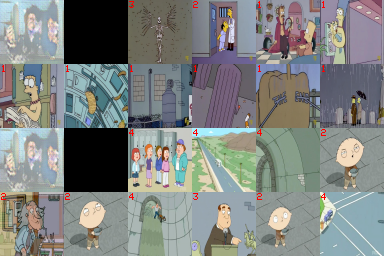}}{{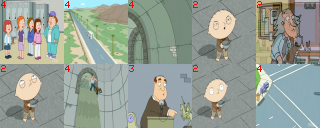}}{{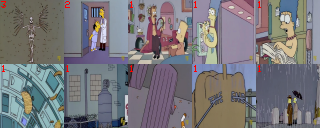}}{{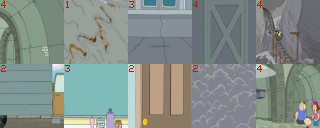}}{{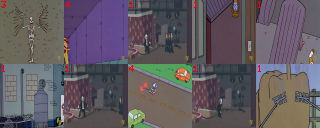}} \midrule
												
												\knndaImgsInit{Simpsons}{{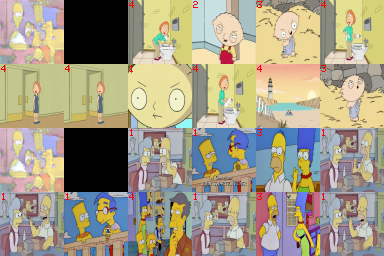}}{{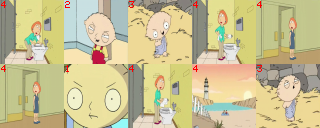}}{{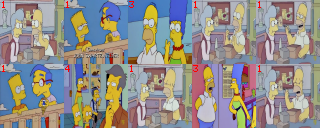}}{{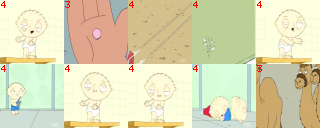}}{{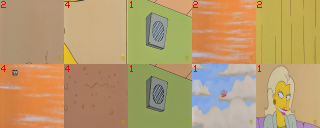}}
												
												\knndaImgs{{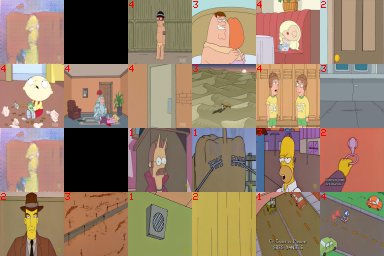}}{{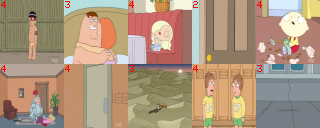}}{{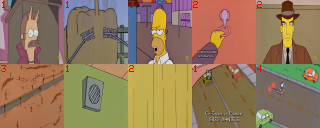}}{{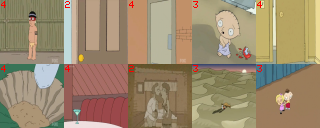}}{{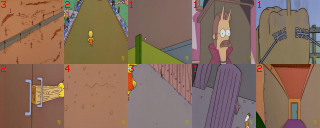}}
												
												\knndaImgs{{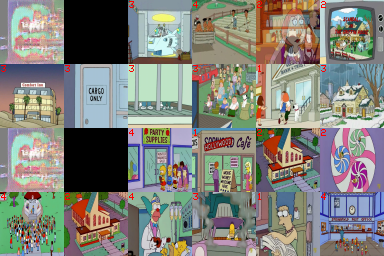}}{{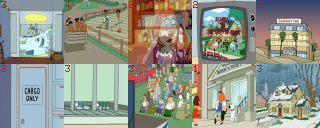}}{{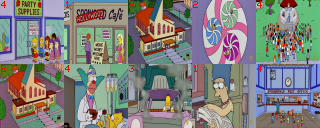}}{{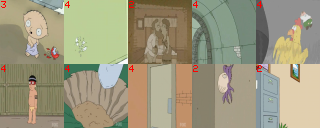}}{{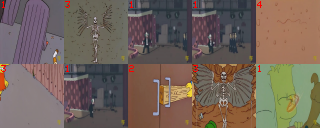}}
												
												\knndaImgsLast{{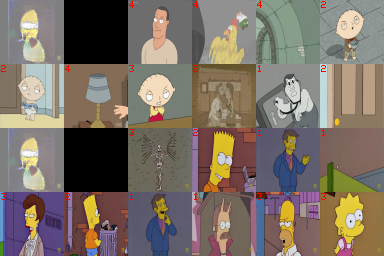}}{{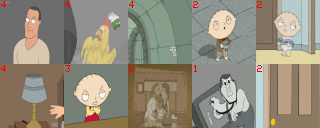}}{{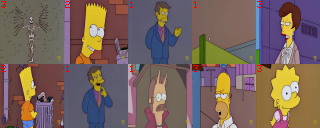}}{{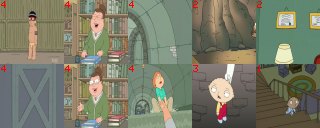}}{{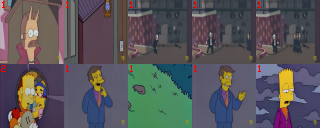}}\midrule
											\end{tabular}
										}}
									\end{center}
									\knncaption{NoClassifierTraining}
									\label{fig:DA_knn_64_v1}
								\end{figure*}

								\xhdr{Domain adaptation variants.}  To better understand the
								importance of different parts of the domain adaptation technique in
								Algorithm 1, we tried various variants, and present results in
								Figure~\ref{fig:DA_all_Samples}. 
								First, our \textbf{NoClassifierTraining} variant
								skips steps 5 and 6  to test if just the shared
								discriminator (without the classifier) is enough to achieve good
								mappings between domains. 
								Figure~\ref{fig:DA_knn_64_v1} shows nearest neighbors for some sample frames generated by this variant of the model.
								For the discriminator similarities, we see much less evidence of semantic correspondence between frames than with the full model,
								although there is some in terms of simple features like overall color.
								% For the discriminator features, we see that in some cases the
								%technique still finds good correspondences, for example between
								%characters and between scenes in the first and third column.  
								%In some cases we can find correspondence
								%between characters as fourth and seventh example while the third
								%example shows correspondence between scenes. % Examining the results
								%shown in Figure \ref{fig:DA_knn_64_v1} we can say that the model
								%didn't find very high level semantic alignments, but the main theme
								%of alignments is domain colors in each domain. In case of the
								%\textit{discriminator similarity}, the system could find good
								%alignments as: \begin{enumerate} 
								%
								%\item The first row can represent
								%alignment of dominant color (yellow). This alignment is represented
								%as images of \freferd{2}{2} and \freferd{2}{3} in Family Guy images
								%matches with images \sreferd{1}{2}, \sreferd{1}{3}, \sreferd{2}{4}
								%and \sreferd{2}{5} in Simpson.
								%
								For example, row 1 has a yellow theme (e.g.\ images 1\freferd{2}{2}
								and 1\freferd{2}{3} with 1\sreferd{1}{2}, 1\sreferd{1}{3},
								1\sreferd{2}{4}, and 1\sreferd{2}{5}), row 2 has a violet theme 
								(images 2\freferd{1}{1}--2\freferd{1}{3} and 2\freferd{1}{5}--2\freferd{2}{1} with 2\sreferd{1}{1}, 2\sreferd{2}{2}, and 2\sreferd{2}{5}),
								rows 4 and 8 features dark gray backgrounds, and row 5 has a combination of yellow and blue colors.
								The
								classifier similarities are not useful here, of course, since they
								have not been trained, and resemble clusterings of random vectors.
								%Also using \textit{Classifier similarity} the results are shown in figure \ref{fig:DA_knn_64_v1}. These results resemble the results of $k$-nearest neighborhood results of random vector since there is no classifier training in this stage. The retrieved samples represents different and and related colorful backgrounds in both domains. 
								%%%%%%%%%%%%%%%%%%%%%%%%%%%%%%%%%%%%%%%%%%%%%%%%%%%%%
								\setcounter{frameno}{0}
								\begin{figure*}
									\begin{center}
										{\footnotesize{
												\begin{tabular}{@{}p{12pt}@{}p{1.55cm}@{\,\,}|@{\,\,}p{3.6cm}@{\,\,}|@{\,\,}p{3.6cm}@{\,\,}|@{\,\,}p{3.6cm}@{\,\,}|@{\,\,}p{3.6cm}@{}}
													\knnheading
													
													\knndaImgsInit{Family Guy}{{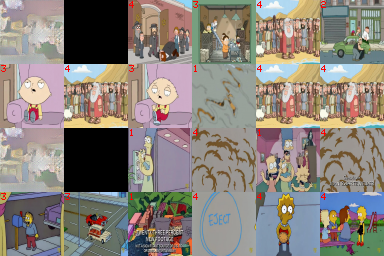}}{{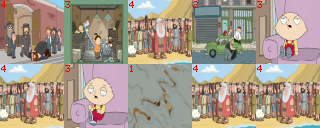}}{{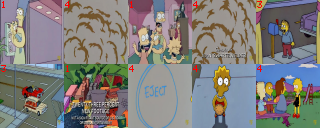}}{{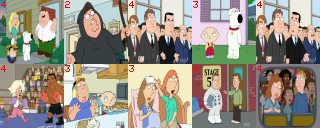}}{{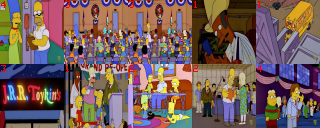}}
													
													\knndaImgs{{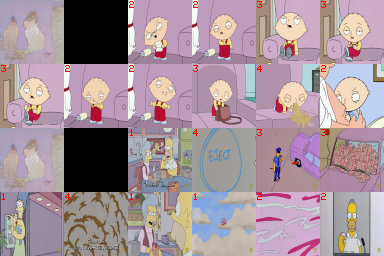}}{{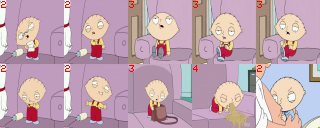}}{{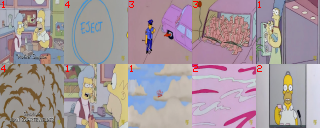}}{{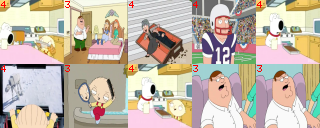}}{{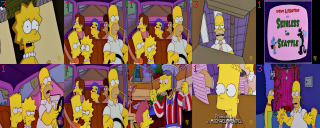}}
													
													\knndaImgs{{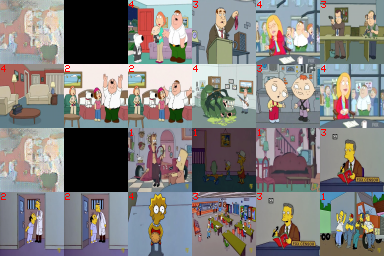}}{{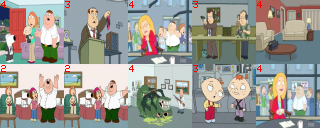}}{{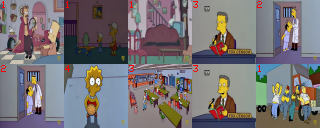}}{{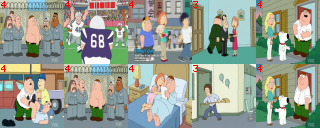}}{{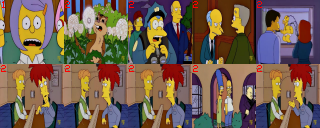}}
													
													\knndaImgsLast{{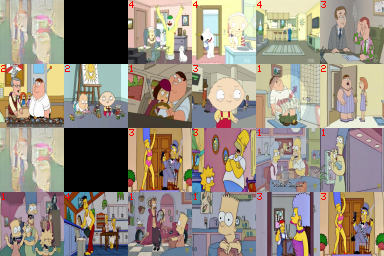}}{{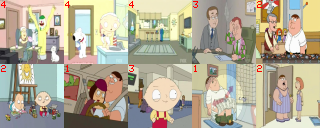}}{{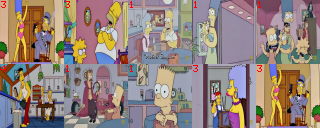}}{{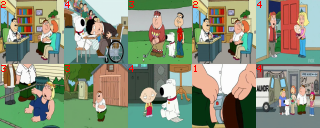}}{{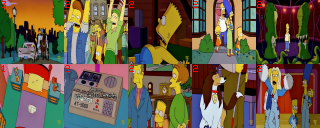}}\midrule
													
													\knndaImgsInit{Simpsons}{{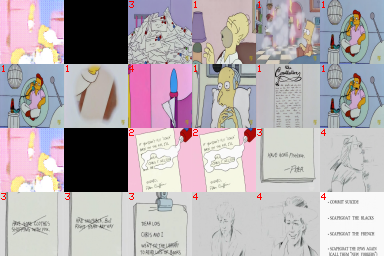}}{{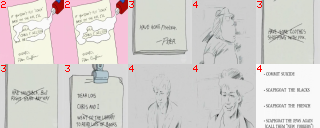}}{{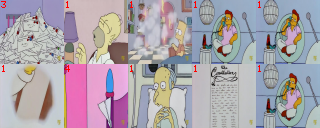}}{{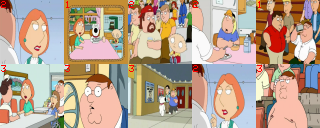}}{{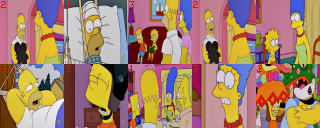}} 
													
													\knndaImgs{{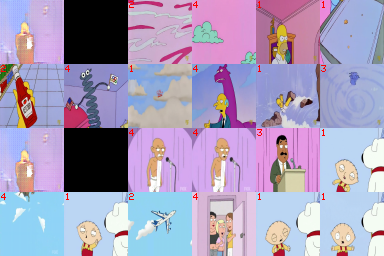}}{{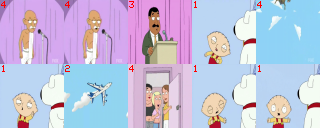}}{{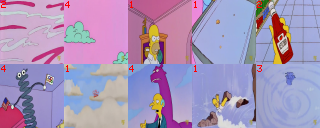}}{{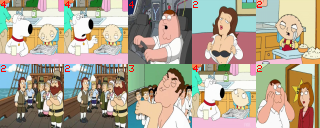}}{{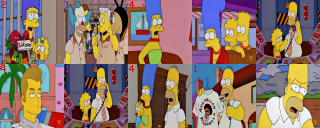}}
													
													\knndaImgs{{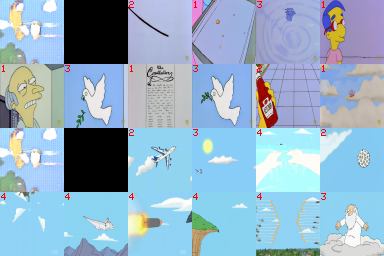}}{{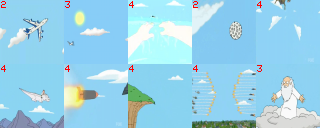}}{{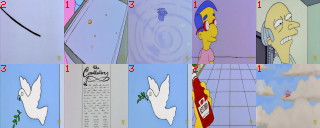}}{{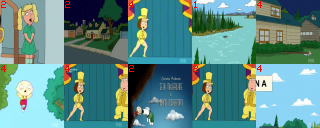}}{{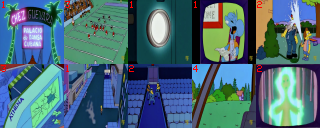}}  
													
													\knndaImgsLast{{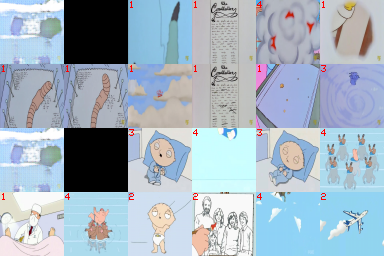}}{{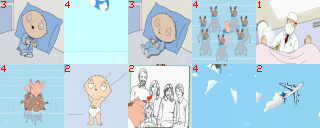}}{{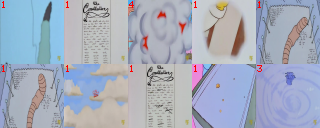}}{{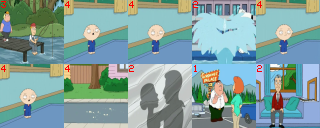}}{{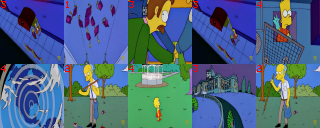}}  \midrule
												\end{tabular}
											}}
										\end{center}
										\knncaption{NoFakeClassifierTraining}
										\label{fig:DA_knn_64_v2}
									\end{figure*}
									
									\textbf{NoFakeClassifierTraining} includes step 5 but skips step 6, so that the
									classifier is trained only with real images, and results are shown in
									Figure~\ref{fig:DA_knn_64_v2}. 
									These results suggest the technique has once again found some higher-level semantic alignments
									in both similarity spaces, including the clouds of dust or smoke in images 1\freferd{2}{3}, 1\sreferd{1}{2}, and 1\sreferd{1}{4}, the large groups of people in 1\freferd{1}{1}, 1\freferd{2}{1}, 1\freferd{2}{4} and 1\freferd{2}{5} with 1\sreferd{2}{5}, 
									the paper documents in images 5\freferd{1}{1}--5\freferd{2}{5} with 5\sreferd{1}{1}, 5\sreferd{2}{1}, and 5\sreferd{2}{4}, the single characters against a blue background in row 6, and the sky-colored background with white foreground objects like clouds in rows 7 and 8.
									Under classifier similarities, we also see some semantic themes,
									including several rows that seem to be cuing on certain facial
									reactions (e.g.\ images 5\freferc{1}{1}, 5\freferc{2}{2},
									5\freferc{2}{4}, and 5\freferc{2}{5} with 5\sreferc{1}{2},
									5\sreferc{2}{1}, 5\sreferc{2}{2}, and 5\sreferc{2}{4}).
									%% Also using \textit{Classifier similarity} the results are shown in figure \ref{fig:DA_knn_64_v2}, where the system find alignments as :
									%% \begin{enumerate}
									%%      \item The row can represent alignment as large congregation of people. This alignment is represented as images of \freferc{1}{3}, \freferc{1}{5} and \freferc{2}{5} in Family Guy images matches with images \sreferc{1}{2}, \sreferc{1}{3}, \sreferc{2}{4} and \sreferc{2}{5} in Simpson.
									
									%%      \item The seventh row can represent alignment as green and blue scenes with people in some cases. This alignment is represented as images of {i}{j} in Family Guy images matches with images {k}{l} and in Simpson where $i,j,k,l = 1 \hdots 5$.
									
									%%      \item The eight row can represent alignment as greenish scene and a single character . This alignment is represented as images of \freferc{1}{1}, \freferc{2}{4} and \freferc{2}{5} in Family Guy images matches with images \sreferc{2}{2}, \sreferc{2}{3} and \sreferc{2}{4} in Simpson.
									
									%%      \item The other rows can represent alignments in terms of people demonstrating different facial reactions. 
									%% \end{enumerate}
									
									% In the second cell, alignment representing crowd of people. Some
									% alignments represent the dominant color as in third cell in the
									% first row, third cell third row and last row. Some alignments are
									% related to interaction between people as in the cells in the second
									% row }
									%%%%%%%%%%%%%%%%%%%%%%%%%%%%%%%%%%%%%%%%%%%%%%%%%%%%%%%%%%%%
									\setcounter{frameno}{0}
									\begin{figure*}
										\begin{center}
											{\footnotesize{
													\begin{tabular}{@{}p{12pt}@{}p{1.55cm}@{\,\,}|@{\,\,}p{3.6cm}@{\,\,}|@{\,\,}p{3.6cm}@{\,\,}|@{\,\,}p{3.6cm}@{\,\,}|@{\,\,}p{3.6cm}@{}}
														\knnheading
														\knndaImgsInit{Family Guy}{{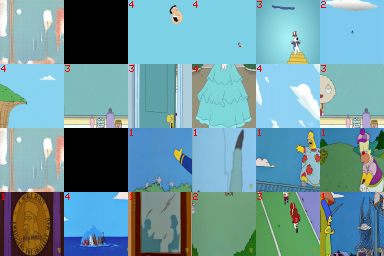}}{{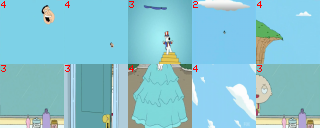}}{{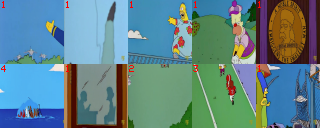}}{{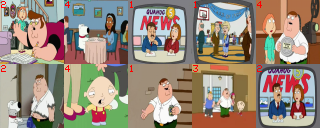}}{{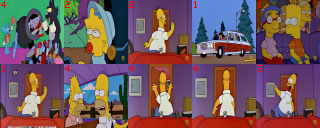}}
														
														\knndaImgs{{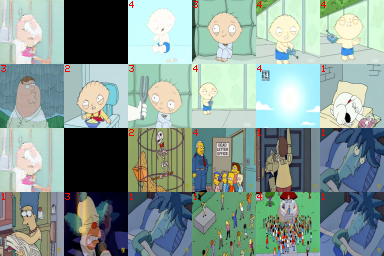}}{{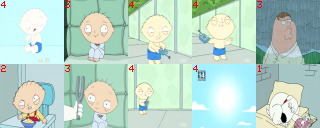}}{{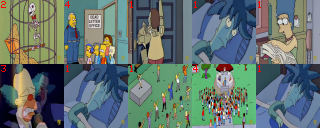}}{{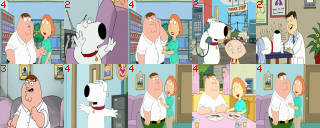}}{{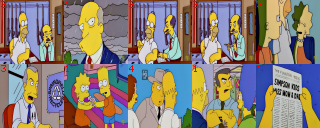}}  
														
														\knndaImgs{{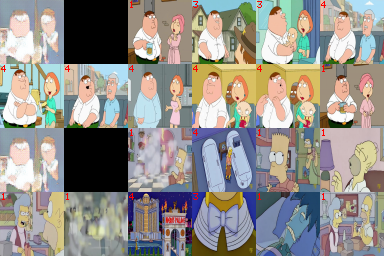}}{{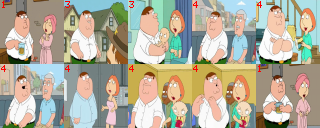}}{{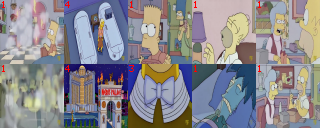}}{{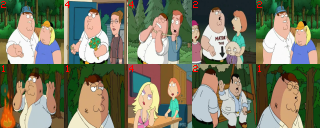}}{{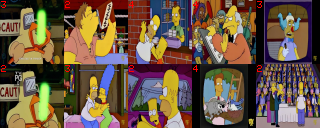}}  
														
														\knndaImgsLast{{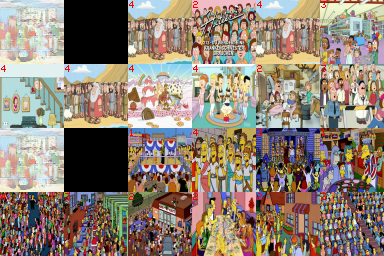}}{{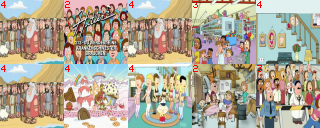}}{{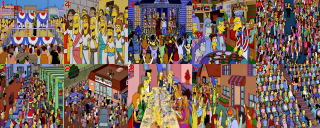}}{{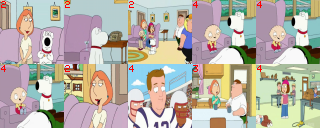}}{{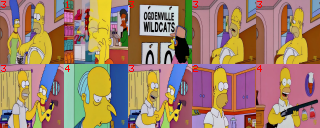}}  \midrule

														\knndaImgsInit{Simpsons}{{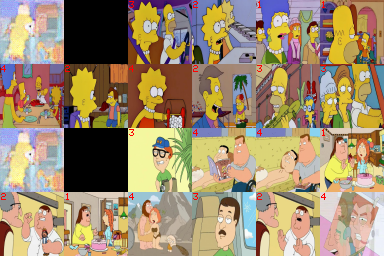}}{{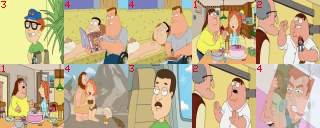}}{{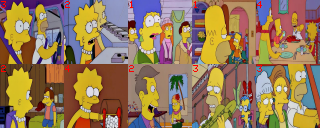}}{{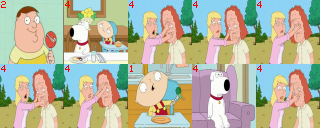}}{{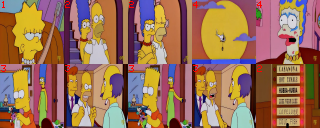}}
														
														\knndaImgs{{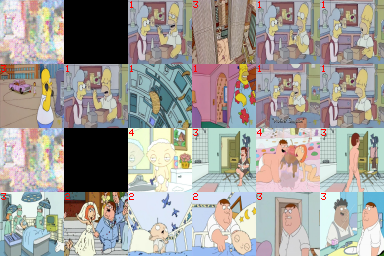}}{{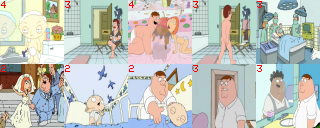}}{{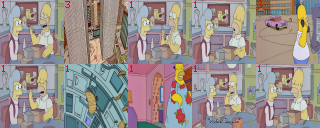}}{{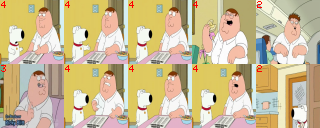}}{{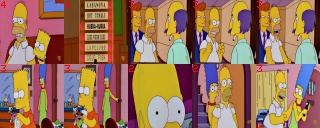}}  
														
														\knndaImgs{{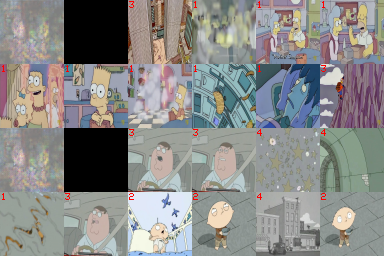}}{{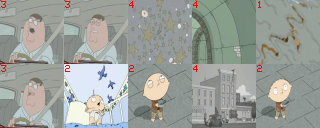}}{{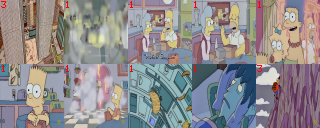}}{{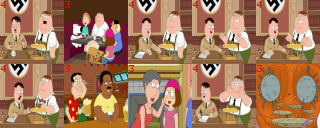}}{{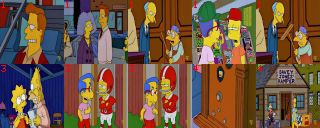}}   
														
														\knndaImgsLast{{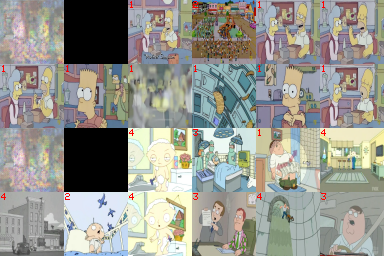}}{{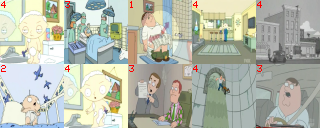}}{{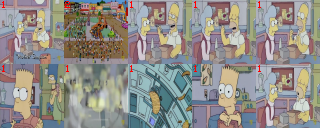}}{{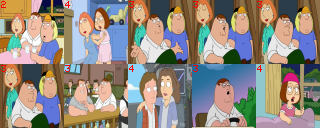}}{{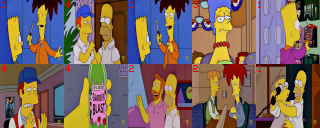}} \midrule 
													\end{tabular}
												}}
											\end{center}
											\knncaption{NoRealClassifierTraining}
											\label{fig:DA_knn_64_v3}
										\end{figure*}
										
										\textbf{NoRealClassifierTraining} skips step 5 but not 6, so that the
										classifier is trained only with synthetic images.  We find that in
										both techniques the model has generated images with good
										correspondence in the two techniques, which suggests that the mapping is
										many to many and not bi-jective, as shown in Figure ~\ref{fig:DA_knn_64_v3}. 
										%It depends on defining the high level
										%semantic meaning the network can find.
										Examples of semantic-level alignments seem to include 
										frames with single humans against a blue background (images 2\freferd{1}{1}--2\freferd{2}{3} with 2\sreferd{1}{3}, 2\sreferd{1}{4}, 2\sreferd{2}{1}, and 2\sreferd{2}{5}), 
										two characters talking (images 3\freferd{1}{1}--\freferd{2}{5} with 3\sreferd{1}{3}, 3\sreferd{1}{5}, and 3\sreferd{2}{5}),
										large crowds of people in row 4, and similar color themes in the remaining rows.
										Classifier similarity results seems to find mostly alignments based on similar character configurations and activities.

										\setcounter{frameno}{0}
										\begin{figure*}
											\begin{center}
												{\footnotesize{
														\begin{tabular}{@{}p{12pt}@{}p{1.55cm}@{\,\,}|@{\,\,}p{3.6cm}@{\,\,}|@{\,\,}p{3.6cm}@{\,\,}|@{\,\,}p{3.6cm}@{\,\,}|@{\,\,}p{3.6cm}@{}}
															\knnheading
															
															\knndaImgsInit{Family Guy}{{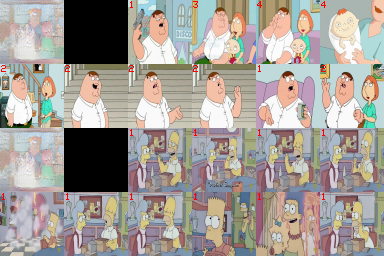}}{{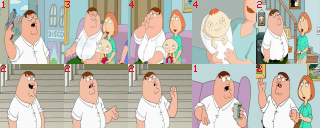}}{{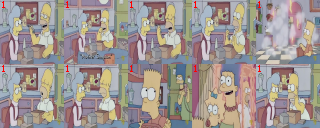}}{{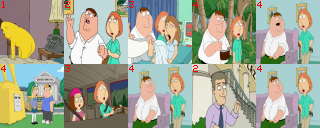}}{{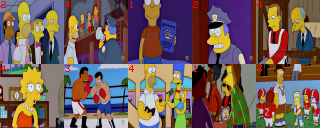}}  
															
															\knndaImgs{{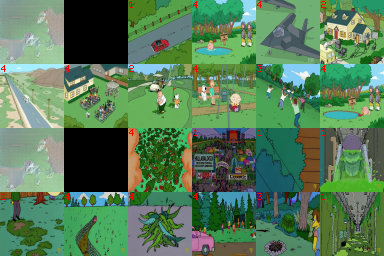}}{{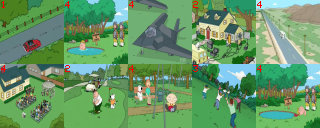}}{{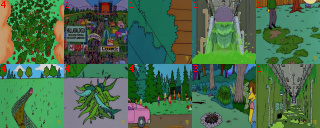}}{{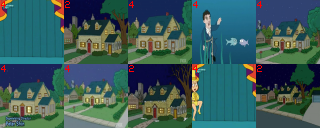}}{{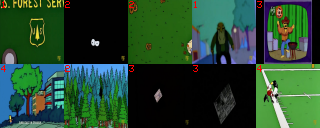}}
															
															\knndaImgs{{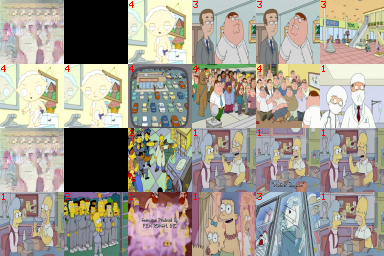}}{{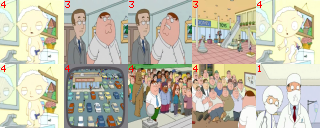}}{{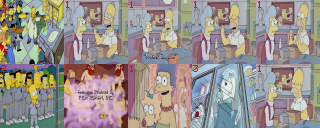}}{{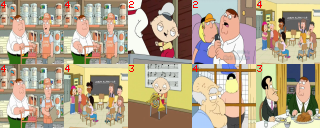}}{{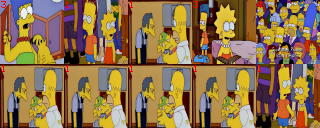}}
															
															\knndaImgsLast{{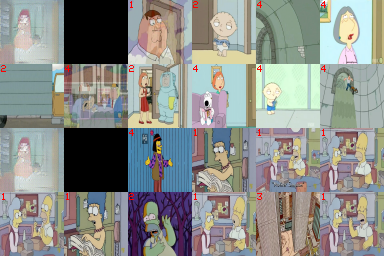}}{{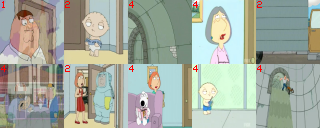}}{{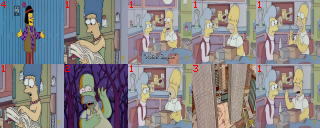}}{{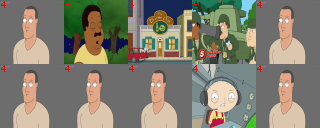}}{{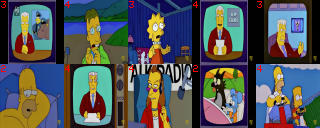}} \midrule
															
															\knndaImgsInit{Simpsons}{{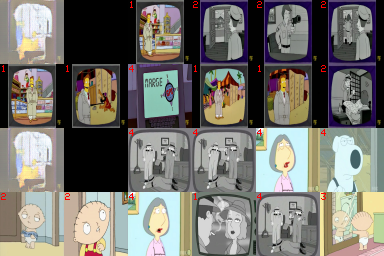}}{{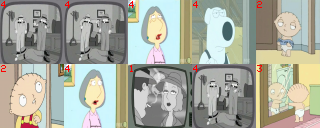}}{{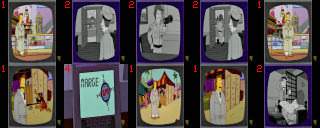}}{{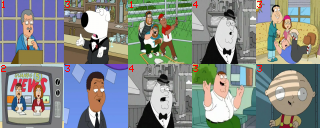}}{{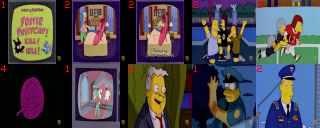}}   
															
															\knndaImgs{{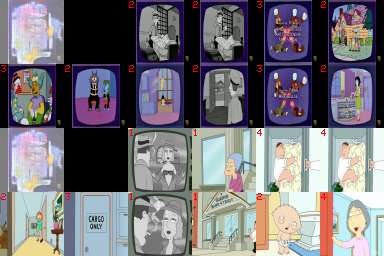}}{{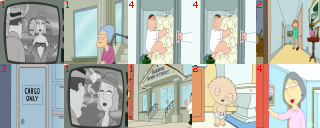}}{{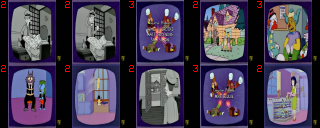}}{{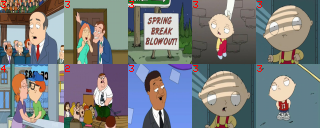}}{{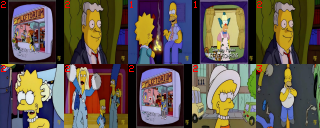}} 
															
															\knndaImgs{{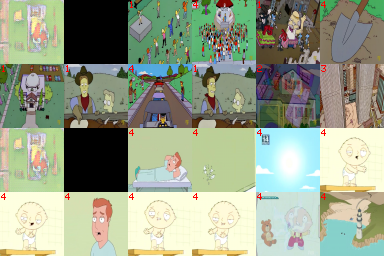}}{{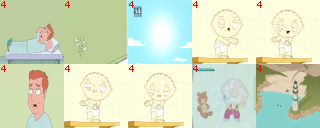}}{{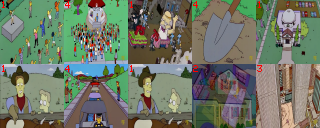}}{{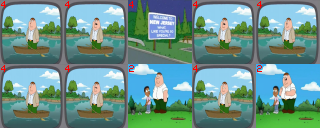}}{{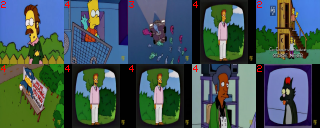}}
															
															\knndaImgsLast{{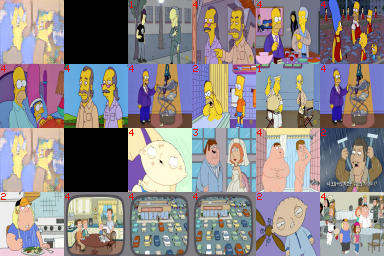}}{{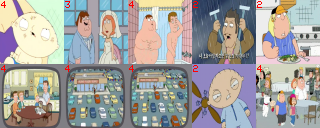}}{{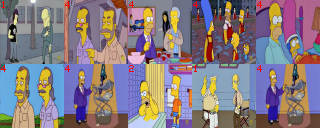}}{{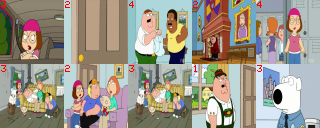}}{{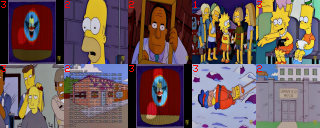}} \midrule 
														\end{tabular}
													}}
												\end{center}
												\knncaption{LazyFakeClassifierTraining}
												\label{fig:DA_knn_64_v4}
											\end{figure*}
											
											Finally, \textbf{LazyFakeClassifierTraining} skips step 6 during the
											initial few iterations, so that the classifier only trains with fake
											images once they start becoming realistic.  
											%We find images of frames
											%(representing TV in the series) are retrieved in both domains, shown%
											%in the fifth and seventh using (discriminator similarity). Also in the
											%sixth example the network finds alignments represented as borders
											%surrounding one character (classifier similarity). Also the third
											%example represents an alignment based on only one character in the
											%scene.  
											Examining the results shown in Figure \ref{fig:DA_knn_64_v4}, we again
											see relatively good high-level semantic alignments, including:
											%
											%we can say that the model can find high level semantic alignments. In
											%case of the \textit{discriminator similarity}, the system could find
											%good alignments as:
											%\begin{enumerate}
											interactions between two main characters (images 1\freferd{1}{5} and 1\freferd{2}{5} with 1\sreferd{1}{1}--1\sreferd{2}{5}),
											blue-green color schemes (row 2), large groups of indoor people (row 3), two characters interacting (images 8\freferd{1}{2} and 8\freferd{1}{3} with 8\sreferd{1}{1}--\sreferd{2}{5}), etc.
											Particularly interesting are rows 5 and 6, where many images correspond with scenes appearing on a TV screen (and thus framed by a rectangular ``viewing window'').
	Correspondences in the classifier similarity space are also readily apparent.
	
	%%%%%%%%%%%%%%%%%%%%%%%
	
\begin{figure}[t]
		\begin{center}
			{\footnotesize{\textsf{
						\begin{tabular}{p{0pt}c@{\,\,}c@{}}
							& Simpsons & Family Guy \\ \toprule
							\multirow{2}{*}[11em]{\begin{sideways}First COGAN iteration\end{sideways}} &
							{\includegraphics[width=0.45\columnwidth]{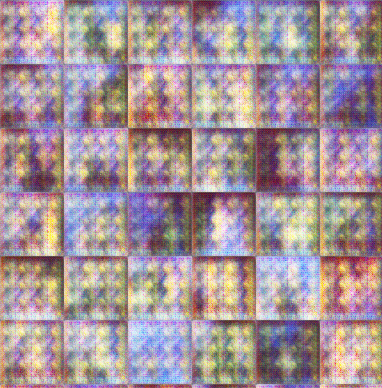}}   &
							{\includegraphics[width=0.45\columnwidth]{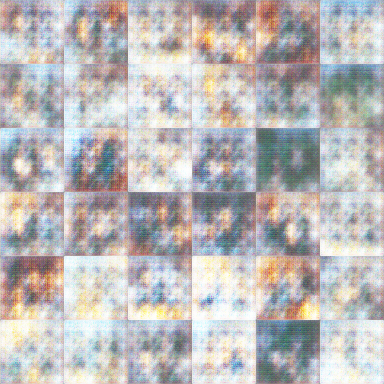}}\\ 
							\multirow{2}{*}[10em]{\begin{sideways}First DA iteration\end{sideways}} &
							{\includegraphics[width=0.45\columnwidth]{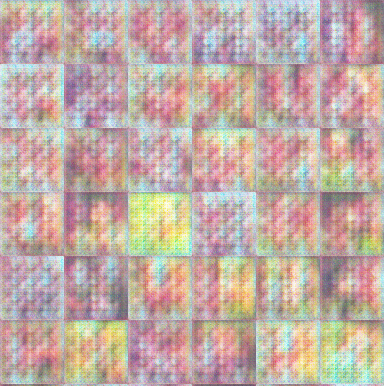}} &
							{\includegraphics[width=0.45\columnwidth]{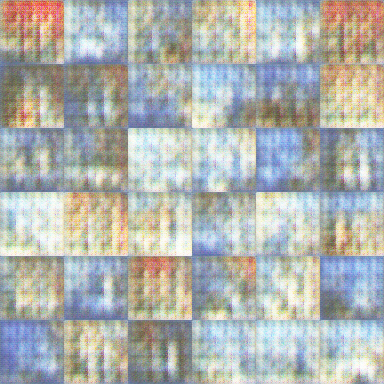}}\\ \bottomrule
							& & \\
							& First iteration & Second iteration\\  \toprule
							\multirow{2}{*}[11em]{\begin{sideways}Simpsons + Family Guy\end{sideways}} &
							{\includegraphics[width=0.45\columnwidth]{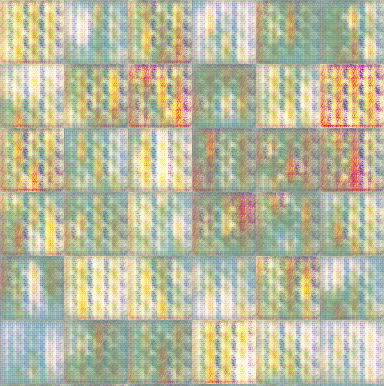}} &
							{\includegraphics[width=0.45\columnwidth]{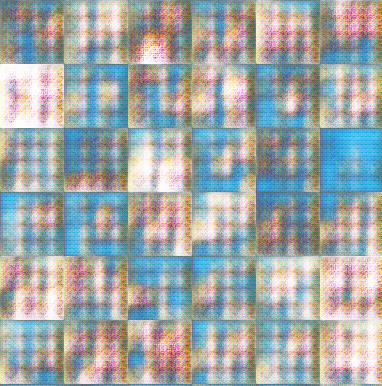}}  \\ \bottomrule
						\end{tabular}
					}}}
				\end{center}
				\vspace{-12pt}
				\caption{Samples generated by various training iterations.}
				\label{fig:iterationsDebug}
			\end{figure}

	We note that when training with both datasets jointly (as opposed to
	training them independently as just one dataset)
	the model manages to
	generate samples that represent
	different color styles of the two series separately, as shown in Figure
	\ref{fig:iterationsDebug}. When trained with a single dataset,
	the model finds it hard to build a joint color space for both
	domains so it alternates between them.
	The first row of Figure
	\ref{fig:iterationsDebug} shows samples generated after the first training
	epoch of Co-GAN model for Family Guy and Simpsons, respectively, while the
	second row shows samples from the domain adaptation
	model. Both the first and second rows show that the network detects
	the difference from the first epoch. The third row of the figure
	shows the first and second epoch of the
	combined dataset image generation. We notice that the model tries to
	find a common color space between the two domains and the results
	change drastically between the two different epochs.

			\newcommand{\RetievalHelper}[2]{
				\setlength\fboxsep{0pt}
				#1 &
				\begin{minipage}{18cm}
					{{{\includegraphics[width=1.\textwidth, height=.15\textwidth]{#2}}}} 
				\end{minipage} &
				\\[-50pt]
				\setcounter{lblno}{0}
				%\labels{a}{b}{c}{d}{e} & \labels{k}{l}{m}{n}{o} & \labels{\textalpha}{\textbeta}{\textgamma}{\textdelta}{\textepsilon} & \labels{\textlambda}{\textmu}{\textnu}{\textxi}{\textomikron} \\[11pt]
				\labels{a}{b}{c}{d}{e}{f}{g}{h}{i}{j}{k}{l}{m}{n}{o}{p}{q}{r}{s}{t} \\[25pt]
				& 
				%\labels{f}{g}{h}{i}{j} &\labels{p}{q}{r}{s}{t} & \labels{\textzeta}{\texteta}{\texttheta}{\textiota}{\textkappa} & \labels{\textpi}{\textrho}{\textsigma}{\texttau}{\textupsilon} \\[-40pt]
				\labels{A}{B}{C}{D}{E}{F}{G}{H}{I}{J}{K}{L}{M}{N}{O}{P}{Q}{R}{S}{T} \\[-40pt]
				& \hspace{2pt}\lblframe & & & \\[32pt]
				%& F1 & F2 & {\multicolumn{1}{r}{{{\textsf{F3}}}}} & F4 & F5  \\[3pt]
			}

			\subsection{An application}
			A potential direct application of our model is in cross-domain image
			retrieval, and in this case, finding semantically-similar episodes
			across two different cartoon series. We test both the discriminator
			and classifier feature-based similarities for
			\textbf{FullDomainAdaptation}. We view each episode as a bag-of-frames, and then given
			an episode in one domain and an episode in another, we calculate a distance
			measure: for every frame in the first episode, we find the closest frame in the second measure, and each frame is allowed to be paired to at most one other frame from the other episode. The distance between the two episodes is the minimum closest frame distance.
			
\begin{figure*}[t]
			\begin{center}
				{\footnotesize{
						\begin{tabular}{@{}p{12pt}c}
							\midrule
							\multirow{8}{*}[0em]{\begin{sideways}{{\textsf{\textbf{Discriminator Similarity}}\,\,\,\,\,\,\,\,\,\,\,\,\,\,\,\,\,}}\end{sideways}} &
							\includegraphics[width=0.9\textwidth, height=0.09\textwidth]{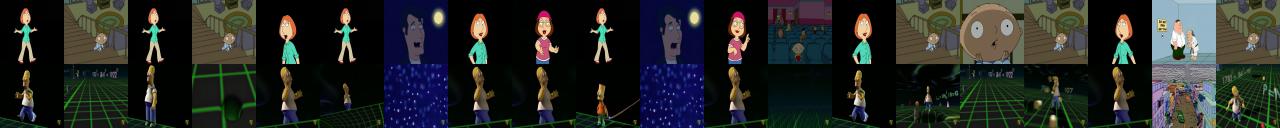} \\ \cline{2-2}
							\\[-8pt]
							& \includegraphics[width=0.9\textwidth, height=0.09\textwidth]{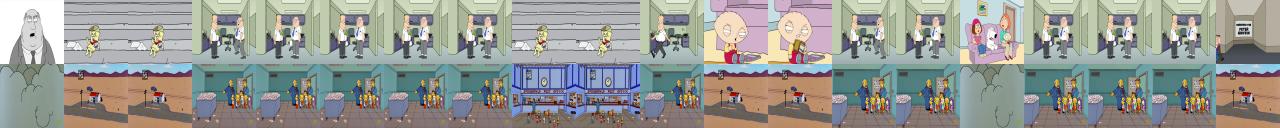}   \\ \cline{2-2} 
							\\[-8pt]
							& \includegraphics[width=0.9\textwidth, height=0.09\textwidth]{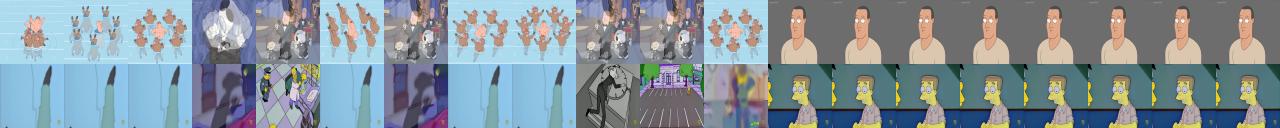}  \\ \cline{2-2}
							\\[-8pt]
							& \includegraphics[width=0.9\textwidth, height=0.09\textwidth]{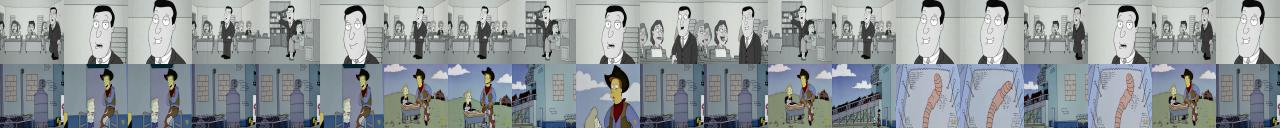} \\ 
							\midrule
							\multirow{7}{*}[0em]{\begin{sideways}{{\textsf{\textbf{Classifier Similarity}}\,\,\,\,\,\,\,\,\,\,\,\,\,\,\,\,\,}}\end{sideways}} &
							\includegraphics[width=0.9\textwidth, height=0.09\textwidth]{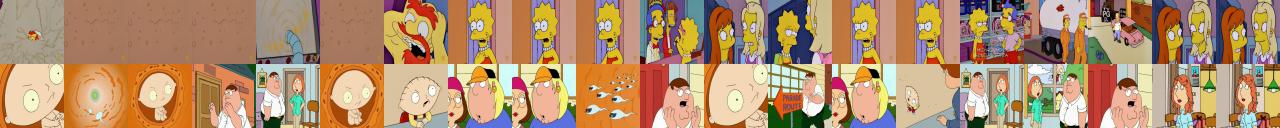} \\\cline{2-2}
							\\[-8pt]
							& \includegraphics[width=0.9\textwidth, height=0.09\textwidth]{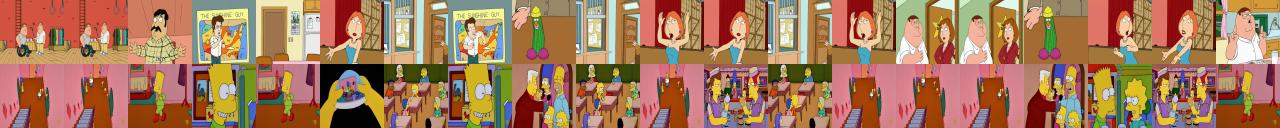} \\\cline{2-2}
							\\[-8pt]
							& \includegraphics[width=0.9\textwidth, height=0.09\textwidth]{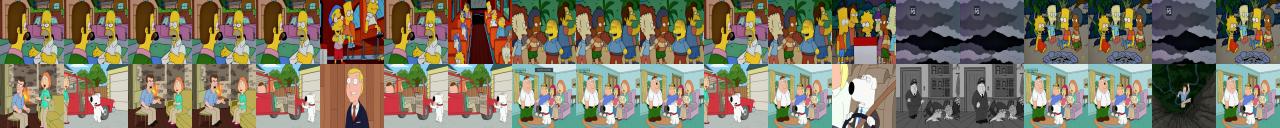} \\\cline{2-2}
							\\[-8pt]
							& \includegraphics[width=0.9\textwidth, height=0.09\textwidth]{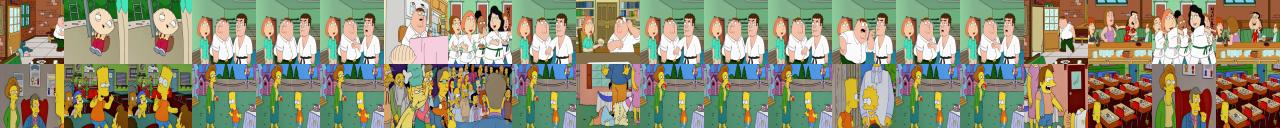}  \\
							\midrule
						\end{tabular}
					}}
				\end{center}
				\vspace{-10pt}
				\caption{Cross-domain similar episode retrieval examples using both discriminator (top) and classifier (bottom) similarity. In each panel, the top row shows a subset of frames from a query video, and the bottom shows the corresponding matched frames from the most similar video in the other domain.}
				\label{fig:retrieval}
			\end{figure*}

			Figure \ref{fig:retrieval} shows four sample retrieval results for
			both the discriminator similarity measure (top) and classifier
			similarity measure (bottom).  Within each result, the top row shows
			$20$ sample frames from a Family Guy query video, and the bottom shows
			the matched frames from the most similar Simpsons episode.  While we
			do not have ground truth to evaluate quantitatively, we see that the
			retrieved results do share similarities in terms of overall scene
			composition.  Examining the results for the discriminator similarity,
			for example, the episodes in the first example both features
			``synthetic''-looking 3d models of characters, while the episodes in
			the second example feature objects against a blue sky and close-ups of
			people. The third example shows indoor rooms from a specific viewpoint,
			and the fourth seems to match black and white scenes in Family Guy with Cowboy scenes in Simpsons.
			Meanwhile, the classifier similarity measure seems to have found simiar
			episodes in terms  of appearance of  main characters and similar background colors.

		%% and the bottom shows The first example First example shows frames of
		%% episode $6$ in the first season of Family Guy that matches episode $6$
		%% in season $7$ of Simpson. The results shows that the two episode are
		%% aligned by displaying a 3D of the main characters together.

		%% Each episode matching result is represented as two rows. Each
		%% row consists of $20$ frames each of Family Guy in the first row and
		%% Simpson in the second row.  The results are described below:
		%% \begin{enumerate}
		%%     \item 
		
		%%     \item Second example shows the match between episode $15$ of second season in Family Guy with episode $5$  of season $7$ in Simpson. The results display that the system matched back and white scenes in Family Guy with Cowboy scene in Simpson. 
		
		%%     \item Third example shows that episode $3$ of third season in Family Guy matches episode $12$ of season $10$ in Simpson. The results display the alignment of indoor cubical rooms between the two domains. 
		
		%%     \item Fourth example shows that episode $19$ of fourth season in Family Guy matches episode $1$ of season $7$ in Simpson. The results shows alignment of one single main character in both domains, also it shows an alignment of main blue sky color in both domains.  
		
		%%     \item Using the classifier similarity we can find that the system find the alignment concerning characters with similar background color
		
		%% \end{enumerate}

\section{Conclusion and Future Work}\label{sec:conclusion}
														
We have studied finding high level semantic mappings between cartoon
frames using GANs, as a first step towards finding general semantic connections
between videos.
We show that this problem is many-to-many (not bi-jective which means that the same scene in one domain can map to different scene in the other domains each mapping can have a high level semantic meaning) and
that some models can find reasonable high-level semantic
alignments between the two domains. Our work also shows, however, that this
is still an open research problem.
Future work should consider the temporal dimension of video, as 
well as adding other modalities like subtitles and audio.
%% In the future we may think about treating the input as video
%% prediction utilizing motion information. Also, we may consider adding
%% some modality like subtitles that can help finding high level semantic
%% correspondence between the two domains.
%
Beyond the insight that our analysis gives about GANs, it also creates the
opportunity for interesting applications in the specific domain of
cartoons.  For example, a common practice among fans is to create
correspondences between live-action movies and TV cartoon series~\cite{SimsonImg1,SimsonImg2}, such as
parodies. Our work raises the intriguing possibility that such
mappings between domains could be created completely automatically by
GAN models. By training on TV series as opposed to individual images,
it may even be possible to sample entirely new story lines, generating
new episodes that fit the stylistic mores of a given series,
completely automatically!

{\small
\bibliographystyle{ieee}
\bibliography{egpaper_final}
}
\end{document}